\documentclass[]{interact}

\usepackage{epstopdf}
\usepackage[caption=false]{
  subfig}

\usepackage{bm}
\usepackage{tabularx}
\usepackage{makecell}
\usepackage{multirow}  
\usepackage{amssymb, amsmath} 

\usepackage{hyperref}
\usepackage{xcolor}
\hypersetup{
    colorlinks=true,
    citecolor=blue,
    linkcolor=blue,
    urlcolor=blue,
}
\usepackage{color}
\usepackage{booktabs} 
\usepackage{fontawesome5}
\usepackage[pagewise]{lineno} 
\definecolor{orcidgreen}{HTML}{A6CE39}
\usepackage[normalem]{ulem}

\usepackage[numbers,sort&compress]{natbib}
\bibpunct[, ]{[}{]}{,}{n}{,}{,}
\renewcommand\bibfont{\fontsize{10}{12}\selectfont}
\makeatletter
\def\NAT@def@citea{\def\@citea{\NAT@separator}}
\makeatother

\theoremstyle{plain}

\theoremstyle{definition}

\theoremstyle{remark}

\begin{document}
\articletype{FULL PAPER}

\title{Multifunctional Locomotion Control of Multi-Jointed BURs
with Swimming and gait Capabilities}

\author{
\name{Takumi Asada \href{http://orcid.org/0009-0001-3050-4635}{\textcolor{orcidgreen}{\faOrcid}}\textsuperscript{a}\thanks{CONTACT Takumi Asada. Email: tasada381@gmail.com}, Hideo Furuhashi \href{http://orcid.org/0000-0003-0015-7203}{\textcolor{orcidgreen}{\faOrcid}}\textsuperscript{b}, Kenta Tabata \href{http://orcid.org/0000-0002-7994-4957}{\textcolor{orcidgreen}{\faOrcid}}\textsuperscript{a}, Renato Miyagusuku \href{http://orcid.org/0000-0003-2471-2187}{\textcolor{orcidgreen}{\faOrcid}}\textsuperscript{a}, and Koichi Ozaki \textsuperscript{a}}
\affil{\textsuperscript{a}Utsunomiya University, 7-1-2 Yoto, Utsunomiya, Tochigi, 321-8585, Japan; \textsuperscript{b}Aichi Institute of Technology, 1247 Yachigusa, Yakusa, Toyota, Aichi, 470-0392, Japan}
}

\maketitle

\begingroup
\renewcommand{\thefootnote}{}
\footnotetext{This is a preprint of an article whose final and definitive form has been published in ADVANCED ROBOTICS 2026, copyright Taylor \& Francis and Robotics Society of Japan, is available online at: \url{https://www.tandfonline.com/doi/full/10.1080/01691864.2026.2728312}; DOI number: 10.1080/01691864.2026.2728312
.}
\endgroup

\begin{abstract}
Multifunctional biomimetic underwater robots (BURs) are capable of conducting underwater tasks suitable for the environment. Combining the characteristics of aquatic organisms enables the swimming and gait locomotion required for underwater exploration. This control mechanism relies on the designer's discretion. This limits the robot's ability to acquire new behavior capabilities. To address these challenges, we propose a mechanism and control system that enables the expression of multifunctional capabilities from the same multi-jointed structure. A mechanism equipped with four leg fins each having four axes is used. Multifunctional control achieves nonlinear behavior based on sensor modalities, rather than relying on predefined conditional switching based on locomotion functions. This system was validated based on multiple sensor modalities and behavior. Utilizing a potential function in multifunctional locomotion control was verified to enable transitions between three behaviors. Implementing the control method as a multifunctional controller is expected to enhance its application in underwater exploration.
\end{abstract}


\section{Introduction}
\begin{figure*}[t]
  \centering
  \resizebox*{14cm}{!}{\includegraphics{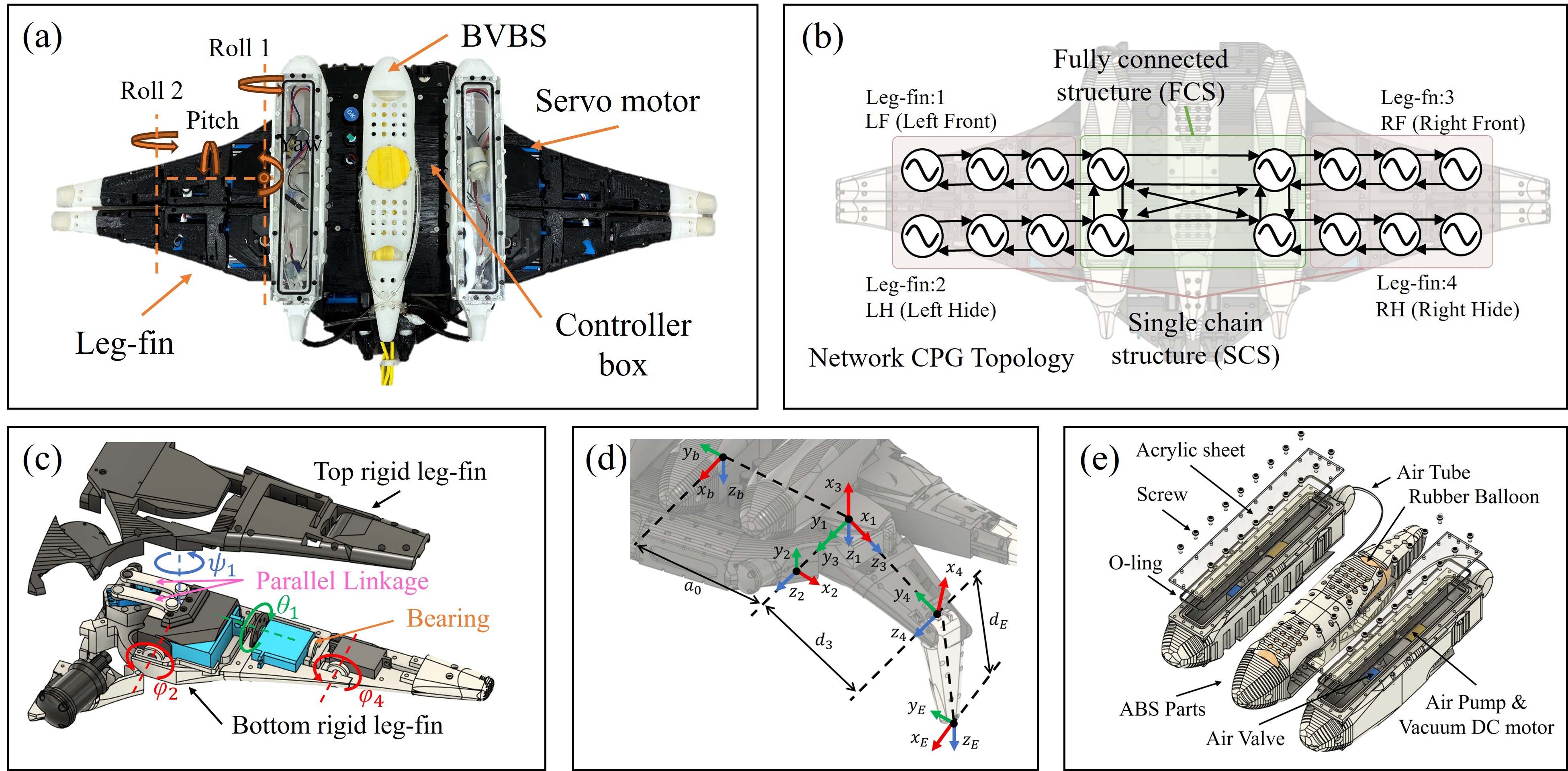}}\hspace{5pt}
  \caption{Prototype of the manta ray robot. (a) overview of the manta ray robot, (b) CPG topological network, (c) mechanical of the leg-fins, (d) DH-parameter of the robot, (e) BVBS mechanical diagram.}
  \label{fig:Fig1}
\end{figure*}
BURs can achieve highly efficient swimming by utilizing the characteristics of the organisms they mimic. These robots are developed for underwater exploration applications \citep{li2024current,cui2023review}, but the exploration environment is limited by the organism they mimic. Only performs the movements of the mimicked organism. Thruster propulsion robots like remotely operated vehicles (ROVs) have a wide exploration range. The visibility is significantly reduced near the seafloor due to the upwelling of sediment \citep{liu2024maneuverable}. By overcoming the challenges of BUR and ROV, the exploration capabilities of underwater robots can be enhanced. Clarification of the mechanisms enabling multifunctionality and their control systems can particularly expand the scope of application for BUR underwater exploration. Multi-jointed structures enable the expression of diverse morphologies using biological characteristics. However, the emergence of multifunctional mechanisms still largely depends on the designers' discretion. Therefore, we propose an underwater robotic mechanism and its control system that emerges with locomotion distinct from conventional biological functions. The fin structure is regarded as a multi-jointed structure and controlled in a distributed system. This concept enables the generation of different locomotion, such as swimming and gait, from the same multi-joint and control network structure. This study focuses on the multi-functional features of BUR, and its structural characteristics and control system.

Multimodal and multifunctional BURs are attracting attention as an extension of underwater exploration. The multimodalization based on the Body and/or Caudal Fin locomotion (BCF) type \citep{sfakiotakis2002review} achieves swimming, floating, and diving through sensory feedback \citep{wang2014cpg, zhang2021design}. Target tasks enable adaptive swimming by changing central pattern generator (CPG) parameters. Multimodal capabilities are achieved for the single function of swimming. A multifunctional BCF robot has also been proposed that achieves both swimming and crawling using three oscillators: two flippers on the left and right and a caudal fin \citep{crespi2008controlling}. Multifunctional locomotion is generated by selecting CPG parameters based on sensor values. This is achieved by pre-selecting unique biological structures capable of exhibiting swimming and gait behaviors. Snake-like robots such as Salamandra and Interplay are also equipped with multifunctional swimming and gait capabilities in advance \citep{crespi2013salamandra, yasui2019decoding}. The drive units exist for both the legs and the whole body, thus enabling adaptive switching of locomotion. A multifunctional median and/or pectoral fin locomotion (MPF) type robot achieve swimming and gai functions using its left and right pectoral fins. Mimicking crabs and turtles achieve multifunctionality through variations in legged movement, material properties, and stiffness \citep{chen2022study, yoo2016design, baines2022multi}. This requires specialized mechanisms, with each function implemented as independent control. To achieve multifunctionality in BURs, each joint is predefined with swimming and gai functions. On the other hand, a multifunctional robot is limited by the imitation target. There are also proposed methods enabling multifunctionality without dependence on the biological model. The adoption of flexible fin or legged structures enables swimming, walking, and crawling \citep{wu2024underwater, kim2021underwater, kim2024development}. By modifying the CPG parameters of the flexible fin drive unit, different locomotor functions such as swimming, walking, and crawling are achieved. Passive joint design and its control methods have an important role in multimodal locomotion. Also, the addition of a new gait mechanism to the existing BCF enables multifunctionality \citep{yan2020research}. This method enables multifunctionality regardless of the type of BUR. As with rule-based control, additional functions require predetermined configuration. The control algorithm switches between swimming and gait locomotion based on predefined thresholds.

Previous studies have achieved swimming and gait functions for each individual joint through material changes and mechanical separation. The role of the target function is provided from the design process. These emerge mechanisms predefine the conditions of multiple behaviors through mapping based on external sensor values and linear functions. This control mechanism relies on the designer's iscretion. The algorithm becomes complex because the conditions for behavioral transitions also vary depending on the imitation target. Furthermore, the material and mechanical structure are unique to each individual joint. It cannot perform other functions beyond its target movement. This is a limitation on the potential for robots to emerge with new capabilities. To overcome these challenges, we propose a mechanism and control system that enables the expression of multifunctional swimming and gait capabilities from the same multi jointed structure. Reproduction of locomotion functions from the same multi-joint structure can eliminate the need of separating or adding functions during the design phase. To investigate the mechanism of expression, we propose a mechanism that utilizes changes in potential functions and phase dynamics to generate behaviors with entirely different locomotion functions. Specifically, swimming and gait locomotion are expressed nonlinearly in response to the sensor modality, rather than being defined by conditional branching based on locomotion function.

The implementation of our ideas is achieved by selecting organisms with large surface areas as driving components. The wider the drive unit, the number of joints and flexibility characteristics can expand the potential for emergence of multifunctionality. Among BURs, the manta ray satisfies these requirements. Manta rays can move highly efficiently with minimal energy consumption due to their large pectoral fins. Previous studies of manta rays have largely focused on the materials and structural characteristics of their fins \citep{xiang2025variable, xiang2025foldable, hao2024bioinspired, liu2022manta}. We define the fin structure of the manta ray robot as a multi-jointed structure to achieve multifunctionality. The control system that smoothly transition between various functions using multiple sensor inputs from the robot, and the experimental results, are reported.

The article is organized as follows. Section \hyperref[sec:Sec2]{2} describes the design and dynamics of the multi-joint manta ray robot. Section \hyperref[sec:Sec3]{3} presents the multifunctional locomotion control system. Section \hyperref[sec:Sec4]{4} discusses the robot experiments results and compares the performance. Finally, section \hyperref[sec:Sec5]{5} offers a summary of this study.

\section{Structural Design of the Manta Ray Robot}
\label{sec:Sec2}
\subsection{Hardware modeling}
\begin{table}[t]
	\caption{Technical specifications of the manta ray robot.}
	\label{table:tbl1}
	\begin{center}
	\begin{tabularx}{\linewidth}{ll}
	Items & Characteristics \\
	\hline\hline
    Size (L $\times$ W $\times$ H) & 0.38 m $\times$ 0.85 m $\times$ 0.14 m \\  \hline
    Total Mass & 8.4 kg \\ \hline
    Fin arrangement & fins(yaw, roll, pitch, and roll) $\times$ 4 \\ \hline
    Control mode & Wired control \\ \hline
    Controller & Jetson Nano B01, Teensy 4.1 \\ \hline
    Power Supply & 5000 mAh, 7.4V, LiPo battery $\times$ 2\\ \hline
    Operation time & Approx. 1.0h \\
    \hline
	\end{tabularx}
	\end{center}
\end{table}
The multi-jointed BUR used in the experiment to perform swimming and gait are shown in Fig. \hyperref[fig:Fig1]{1}(a), with its specifications shown in Table \hyperref[table:tbl1]{1}. The total weight of the model is 8.4 kg, and the dimensions of the robot are 0.38 m in length, 0.85 m in width, and 0.14 m in height. The shape of the pectoral fins on the body was determined using a third-order polynomial approximation \citep{luo2019parametric}. The body section was designed as NACA0022, and the pectoral fin section as NACA0012. To generate different locomotion from the same multi-joint structure, a two-layer CPG structure is used as shown in Fig. \hyperref[fig:Fig1]{1}(b). Each joint is equipped with four leg-fins, each having four axes: yaw, roll1, pitch, and roll2 in Fig. \hyperref[fig:Fig1]{1}(c). The link relationships are shown in Fig. \hyperref[fig:Fig1]{1}(d) and are listed as Denavit-Hartenberg (DH) parameters in Table \hyperref[table:tbl2]{2}. The homogenous transformation matrix is expressed by Eq. \hyperref[eq:eq1]{(1)}.
\begin{equation}\label{eq:eq1}
  \begin{aligned}
	& ^{i-1}T_i(\theta_i, d_i, a_i, \alpha_i) = R_z(\theta_i) \cdot T_z(d_i) \cdot T_x(a_i) \cdot R_x(\alpha_i), \\
	& ^{i-1}T_i = \begin{bmatrix}
	c\theta_i & -s\theta_i c\alpha_i & s\theta_i s\alpha_i & a_i c\theta_i \\
	s\theta_i & c\theta_i c\alpha_i & -c\theta_i s\alpha_i & a_i s\theta_i \\
	0 & s\alpha_i & c\alpha_i & d_i \\
	0 & 0 & 0 & 1
	\end{bmatrix},
  \end{aligned}
\end{equation}
where \(c\theta_i\) is \(\cos(\theta_i)\), \(s\theta_i\) is \(\sin(\theta_i)\), \(c\alpha_i\) is \(\cos(\alpha_i)\), \(s\alpha_i\) is \(\sin(\alpha_i)\) and $\theta_i$, $d_i$, $a_i$, and  $\alpha_i$ are the joint angle, the link offset, the link length, and the link twist angle, respectively. The constant value $\theta_{init}$ represents the rotation of the center of gravity (CG) about the z-axis.

\begin{table}[t]
	\caption{D-H Parameters of the manta ray robot in NED frame.}
	\label{table:tbl2}
  \begin{center}
	\begin{tabular}{lllll}
	Link & $\theta_i$(deg) & $d_i$(mm) & $a_i$(mm) & $\alpha_i$(rad) \\
	\hline
  \hline
  0 &  $\theta_{init}$ & 0 & $a_0$ & 0 \\
	1 &  $\theta_1+\theta_{\text{offset}}$ & 0 & 0 & -$\frac{\pi}{2}$ \\
	2 & $\theta_2+\frac{\pi}{2}$ & 0 & 0 & $\frac{\pi}{2}$ \\
	3 & $\theta_3$ & $d_3$ & 0 & -$\frac{\pi}{2}$ \\
	4 & $\theta_4$ & 0 & 0 & $\frac{\pi}{2}$ \\
	E & $\frac{\pi}{2}$ & $d_E$ & 0 & 0 \\
	\hline
	\end{tabular}
  \end{center}
\end{table}

This multi-jointed structure was designed as a lateral leg configuration to achieve an increased stride length and improved gait locomotion performance \citep{asada2024development}. In other words, it has four leg-fins, including the yaw axis. Buoyancy adjustment of the body is performed using the balloon variable buoyancy system (BVBS) in Fig. \hyperref[fig:Fig1]{1}(e). The body is designed to be heavier than neutral buoyancy by default, enabling depth adjustment and underwater locomotion through the inflation and deflation of the BVBS. gait locomotion requires torque to reliably push off surfaces like the seabed and ground contact. To enable the expression of different locomotor functions such as swimming and gait in the water, a buoyancy control system is installed on the upper part of the body.

\subsection{Dynamics of the manta ray robot}
\begin{figure}[t] \label{fig:Fig2}
  \centering
  \resizebox*{8cm}{!}{\includegraphics{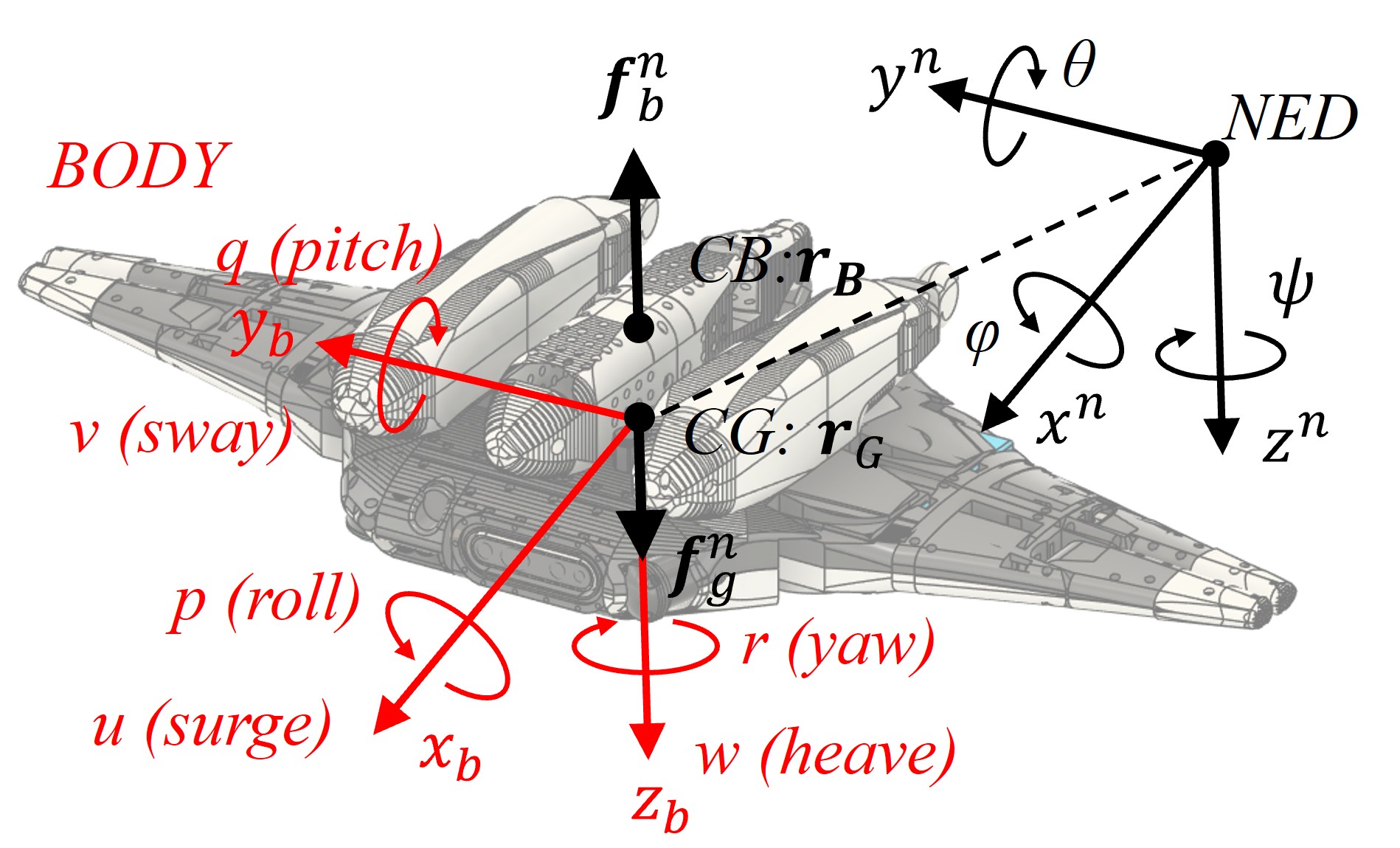}}
  \caption{Coordinate system of the manta ray robot in the NED frame.}
\end{figure}

The multi-jointed BUR of manta ray robot generates thrust force through the swing of its leg-fins. The forces acting due to multi-joint motion are determined by solving the robot's kinematics. The motion of an underwater robot in three-dimensional space is defined by two coordinate systems: the earth-fixed and the body-fixed coordinate system. The earth-fixed frame $\bm{\eta} = [x^n, y^n, z^n, \varphi, \theta, \psi]^T$ is defined by the $x$-, $y$-, and $z$-positions and roll, pitch, and yaw rotation. For underwater robots, this refers to the movements of surge, sway, and heave, and the rotation of heel, trim, and heading angle. The body-fixed frame $\bm{\nu} = [u, v, w, p, q, r]^T$ is defined by the $x$-, $y$-, and $z$-direction velocities, and the roll, pitch, and yaw angular velocities respectively. Figure \hyperref[fig:Fig2]{2} shows the coordinate system of the manta ray robot based on the Fossen model \citep{Fossen2021}.

In this study, we consider the 5 DOF (surge, sway, heave, pitch, and yaw) motions and formulate the dynamic model of the multi-jointed BUR as shown in Eq. \hyperref[eq:eq2]{(2)}:
\begin{equation}\label{eq:eq2}
  \begin{aligned}
  &\bm{M}\bm{\dot{\nu}} + \bm{C}(\bm{\nu})\bm{\nu} + \bm{D}(\bm{\nu})\bm{\nu} + \bm{g}(\bm{\eta}) = \bm{\tau}, \\
  \bm{M} &=
  \resizebox{0.45\textwidth}{!}{$
  \begin{bmatrix}
  m - X_{\dot{u}} & 0 & 0 & 0 & 0 \\
  0 & m - Y_{\dot{v}} & 0 & 0 & 0 \\
  0 & 0 & m - Z_{\dot{w}} & 0 & 0 \\
  0 & 0 & 0 & I_{y} - M_{\dot{q}} & 0 \\
  0 & 0 & 0 & 0 & I_{z} - N_{\dot{r}}
  \end{bmatrix}
  $}, \\
  \bm{D} &=
  \resizebox{0.65\textwidth}{!}{$
    \begin{bmatrix}
    - X_{u}-X_{u|u|}|u| & 0 & 0 & 0 & 0 \\
    0 & - Y_{v} - Y_{v|v|}|v| & 0 & 0 & 0 \\
    0 & 0 & - Z_{w} - Z_{w|w|}|w| & 0 & 0 \\
	0 & 0 & 0 &  - M_{q} - M_{q|q|}|q| & 0 \\
    0 & 0 & 0 & 0 & - N_{r} - N_{r|r|}|r|
  \end{bmatrix}
  $}, \\
\end{aligned}
\end{equation}
where $\bm{M}$ $\in$ $\bm{R^{n \times n}}$ is a symmetric inertia matrix including added mass, $\bm{D(\nu)}$ $\in$ $\bm{R^{n \times n}}$ is a damping matrix, $\bm{g(\eta)}$ $\in$ $\bm{R^{n}}$ is a gravity matrix, $\bm{\tau}$ $\in$ $\bm{R^{n}}$ is the force and moment matrix, respectively, with $n = 5$. The Coriolis and centripetal terms, which include the additional mass $\bm{C(\nu)}$ $\in$ $\bm{R^{n \times n}}$, are small and ignorable in the BUR under slow speed.

The gravitational force $\bm{f}_{g}^n = [0, 0, W]^T$ acts through the CG defined by the vector $\bm{r}_{G} = [x_G, y_G, z_G]^T$ relative to the coordinate origin (CO). The buoyancy force $\bm{f}_{b}^n = [0, 0, -B]^T$ acts through the center of buoyancy (CB) defined by the vector $\bm{r}_{B} = [x_B, y_B, z_B]^T$ relative to the CO. The forces and moments acting about the CO with 5 DOF of gravity and buoyancy are given by Eq. \hyperref[eq:eq3]{(3)}:
\begin{equation}\label{eq:eq3}
  g(\bm{\eta}) =
  \resizebox{0.35\textwidth}{!}{$
  \begin{bmatrix}
    (W - B) s\theta \\
    -(W - B) c\theta s\phi \\
    -(W-  B) c\theta c\phi \\
    (z_G W - z_B B)s\theta + (x_G W - x_B B)c\theta c\phi \\
    -(x_G W - x_B B)c\theta s\phi - (y_G W - y_B B)s\theta
  \end{bmatrix}.
  $}
\end{equation}
Added mass coefficient included in the inertia matrix $\bm{M_a} = -\mathrm{diag}[X_{\dot{u}}, Y_{\dot{v}}, Z_{\dot{w}}, M_{\dot{q}}, N_{\dot{r}}]$ represents the hydrodynamic forces and moment coefficients acting on a body under acceleration. Because the multi-jointed BUR consists of the main body and both left and right leg-fins, the total added mass is calculated as the sum of the main body and the leg-fins. The main body is approximated as a two-dimensional ellipse as both a longitudinal and transverse section. This added mass is determined using strip theory \citep{ newman2018marine}. The added mass of the leg fin section is calculated as a flat plate \citep{ev1989principles}.

\begin{figure*}[t] \label{fig:Fig3}
  \centering
  \resizebox*{14cm}{!}{\includegraphics{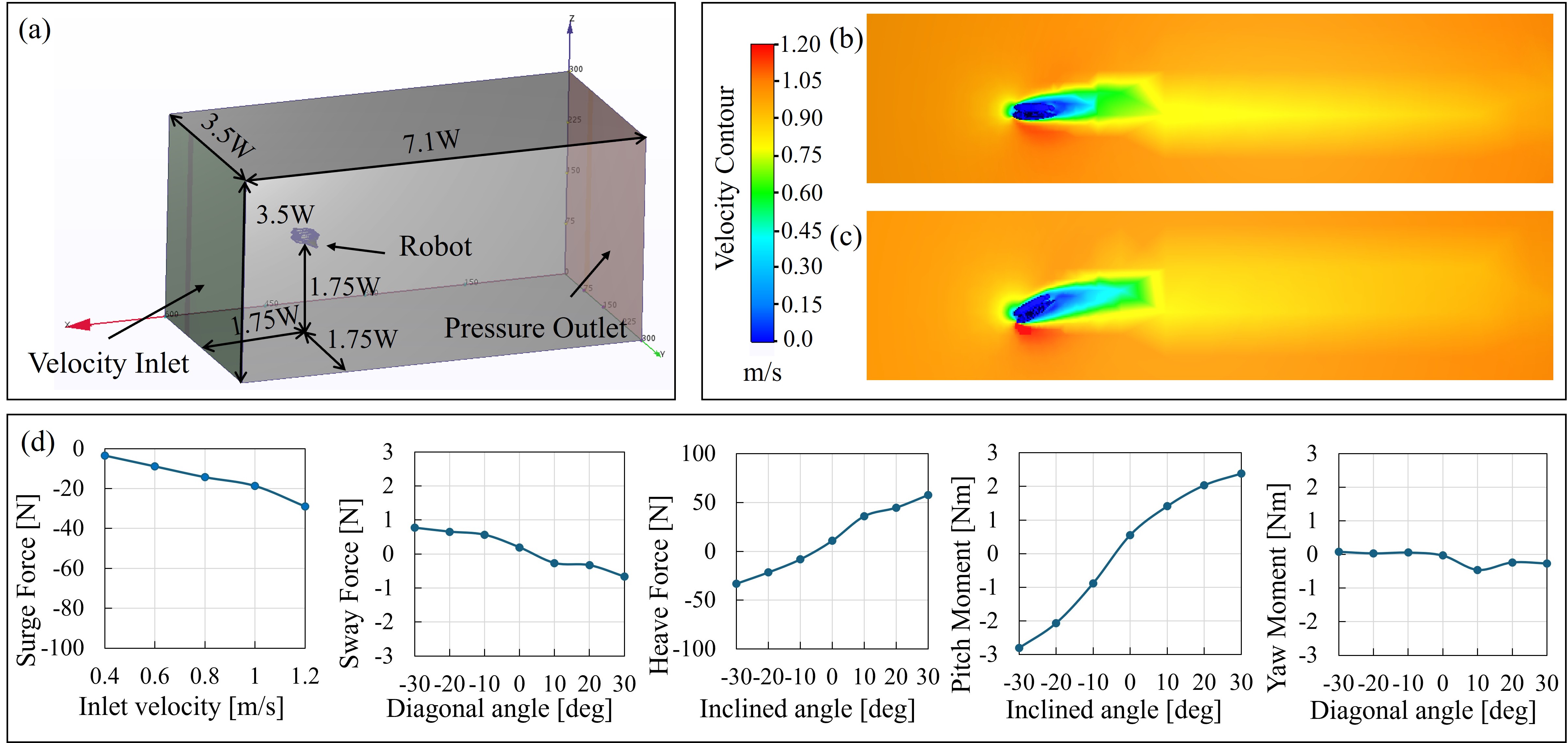}}
  \caption{Computational domain and CFD simulation result, (a) computational domain and boundary conditions, (b) analysis results of pitching angle 0\textdegree, (c) analysis results of pitching angle 30\textdegree, (d) forces and moments result obtained by the CFD simulation.}
\end{figure*}

The linear and nonlinear damping coefficients are determined by solving the overdetermined system using the least squares method \citep{go2019hydrodynamic}. To identify these coefficients, the hydrodynamic forces and moments are expressed as a linear combination of velocity-dependent terms. The damping coefficients of surge $X_u$ and heave $Z_w$ can be calculated using the damping force model and the solution to the matrix equation given by Eq. \hyperref[eq:eq4]{(4)}:
\begin{equation}\label{eq:eq4}
  \begin{aligned}
      \begin{pmatrix} X_1 \\ X_2 \\ \vdots \\ X_n \end{pmatrix}
        &= \begin{pmatrix}
        u_1 & u_1|u_1| \\
        u_2 & u_2|u_2| \\
        \vdots & \vdots \\
        u_n & u_n|u_n|
        \end{pmatrix}
        \begin{pmatrix} X_u \\ X_{u|u|} \end{pmatrix}, \\
    \mathbf{h} &= \left(\mathbf{A}^T \mathbf{A}\right)^{-1} \mathbf{A}^T \mathbf{b}. \\
  \end{aligned}
\end{equation}

The damping coefficients $Y_v$, $M_q$, and $N_r$ are similarly determined by solving an overdetermined system. The damping coefficients for multi-joint BURs are calculated using analysis results from the fluid dynamics software Autodesk CFD. Figure \hyperref[fig:Fig3]{3} shows the computational domain and CFD simulation results. The green front surface and the red rear surface were defined as the velocity inlet surface and the zero-pressure outlet surface, respectively. The simplified robot model was positioned 1.75 $\times$ width (W) away from the velocity inlet surface. The length, width, and height of the computational domain were set to 7.1W, 3.5W, and 3.5W, respectively. The standard $k-\epsilon$ method is used for the turbulence model, and the standard wall function method is applied to the region near the wall. Each damping coefficient is measured using a static analysis method at inflow velocities of 0.4 to 1.2 [m/s], pitching (inclined) and yawing (diagonal) angles of -30 to 30 \textdegree, respectively. The measured values obtained from each simulation case are used in the overdetermined system in Eq. \hyperref[eq:eq4]{(4)}.

\section{Multifunctional Locomotion Control}
\label{sec:Sec3}
To achieve multifunctional locomotion control, this paper proposes a control system consisting of a motor execution system and a behavior transition system.
\subsection{Motor execution system}
Motion execution systems are based on CPG models to realize swimming and gait locomotion from the same multi-joint structure using a single algorithm given by Eq. \hyperref[eq:eq5]{(5)}:
\begin{equation}
  \begin{aligned}
    \dot{\phi_i} &= 2 \pi f_i + \sum_j \omega_{i,j} sin(\phi_j - \phi_i - \Delta \varphi_{ij}), \\
    \ddot{r_i} &= a_i (\frac{a_i}{4} (R_i - r_i) - \dot{r_i}), \\
    \ddot{\chi_i} &= b_i (\frac{b_i}{4} (X_i - \chi_i) - \dot{\chi_i}), \\
    \theta_i &= \chi_i + r_i sin(\phi_i). \\ \label{eq:eq5}
  \end{aligned}
\end{equation}
The parameter of $\phi_i$ , $r_i$ , $\chi_i$ , $\theta_i$ are the phase, amplitude, bias amplitude and output angle of oscillator $i$, respectively. Multifunctional motion is generated from the same multi-joint structure by changing these parameters in response to external sensor modalities. The CPG structure consists of a single chain structure (SCS) in the first layer, connecting each leg-fin to its lateral oscillators. The second layer is a fully connected structure (FCS) that connects the base of each leg-fin to each oscillator as shown in Fig. \hyperref[fig:Fig1]{1}(b). The first layer (SCS) primarily contributes to the amplitude, offset angle, and phase difference of swimming and gait, while the second layer (FCS) contributes to determining the phase difference of locomotion. The target for swimming and gait is defined by the matrices $\bm{R}$, $\bm{X}$, $\bm{\Phi}$, and $\bm{\Phi^{\prime}}$, respectively, as given in Eq. \hyperref[eq:eq6]{(6)} and \hyperref[eq:eq7]{(7)}.
\begin{equation}\label{eq:eq6}
  \begin{aligned}
    \bm{R_{swim}} &=
    \begin{bmatrix}
      0 & \frac{\pi}{4} & \frac{\pi}{4} & \frac{\pi}{4} \\
      0 & \frac{\pi}{4} & \frac{\pi}{4} & \frac{\pi}{4} \\
      0 & \frac{\pi}{4} & \frac{\pi}{4} & \frac{\pi}{4} \\
      0 & \frac{\pi}{4} & \frac{\pi}{4} & \frac{\pi}{4}
    \end{bmatrix},
    \bm{X_{swim}} =
    \begin{bmatrix}
      0 & 0 & 0 & 0 \\
      0 & 0 & 0 & 0 \\
      0 & 0 & 0 & 0 \\
      0 & 0 & 0 & 0 \\
    \end{bmatrix}, \\
    \bm{\Phi_{swim}} &=
    \begin{bmatrix}
      0 & -\pi & 0 & 0 \\
      0 & -\pi & 0 & 0 \\
      0 & -\pi & 0 & 0 \\
      0 & -\pi & 0 & 0
    \end{bmatrix},
    \bm{\Phi^{\prime}_{swim}} =
    \begin{bmatrix}
      0 & -\frac{\pi}{18} & 0 & 0 \\
      \frac{\pi}{18} & 0 & 0 & 0 \\
      0 & 0 & 0 & -\frac{\pi}{18} \\
      0 & 0 & \frac{\pi}{18} & 0
    \end{bmatrix}.
  \end{aligned}
\end{equation}
\begin{equation}\label{eq:eq7}
  \begin{aligned}
    \bm{R_{gait}} &=
    \begin{bmatrix}
      \frac{\pi}{9} & \frac{\pi}{18} & \frac{\pi}{12} & 0 \\
      -\frac{\pi}{9} & \frac{\pi}{18} & \frac{\pi}{12} & 0 \\
      \frac{\pi}{9} & \frac{\pi}{18} & \frac{\pi}{12} & 0 \\
      -\frac{\pi}{9} & \frac{\pi}{18} & \frac{\pi}{12} & 0
    \end{bmatrix},
    \bm{X_{gait}} =
    \begin{bmatrix}
      \frac{\pi}{9} & -\frac{\pi}{6} & -\frac{\pi}{6} & -\frac{5 \pi}{18} \\
      \frac{\pi}{9} & -\frac{\pi}{6} & \frac{\pi}{6} & -\frac{5 \pi}{18} \\
      \frac{\pi}{9} & -\frac{\pi}{6} & -\frac{\pi}{6} & -\frac{5 \pi}{18} \\
      \frac{\pi}{9} & -\frac{\pi}{6} & \frac{\pi}{6} & -\frac{5 \pi}{18} \\
    \end{bmatrix}, \\
    \bm{\Phi_{gait}} &=
    \begin{bmatrix}
      \frac{2 \pi}{3} & -\pi & \pi & 0 \\
      \frac{2 \pi}{3} & -\pi & \pi & 0 \\
      \frac{2 \pi}{3} & -\pi & \pi & 0 \\
      \frac{2 \pi}{3} & -\pi & \pi & 0
    \end{bmatrix},
    \bm{\Phi^{\prime}_{gait}} =
    \begin{bmatrix}
      0 & \pi & \pi & 0 \\
      -\pi & 0 & 0 & -\pi \\
      -\pi & 0 & 0 & -\pi \\
      0 & \pi & \pi & 0
    \end{bmatrix}.
  \end{aligned}
\end{equation}
The columns of $\bm{R}$ and $\bm{X}$ represent the four leg-fins, and the rows represent the settings for each leg-fin oscillator. The element $\bm{\Phi}$ is $\Delta \varphi_{i,i+1}$, denoting the phase difference between adjacent oscillators, where the index $i$ corresponds to the number of oscillators, $i = 1, 2, \ldots, 16$. The $\bm{\Phi^{\prime}}$ element represents the phase difference between oscillators $4k-3$, which determines the phase difference in the second layer FCS. The index $k$ is $k = 1, 2, \ldots, 4$.

Multimodal locomotion behaviors can be generated from the same multi-joint structure by varying the phase difference of the multi-joint structure. The trot gait is activated when $\bm{\Phi^{\prime}_{trot}} = \bm{\Phi^{\prime}_{gait}}$, walk and turning are activated as $\bm{\Phi^{\prime}_{walk}}$ and $\bm{\Phi^{\prime}_{turning}}$ respectively. Lateral gait is achieved by defining the crab gait as $\bm{R_{crab}}$ and $\bm{\Phi_{crab}}$. The other parameters for crab gait are $\bm{X_{crab}} = \bm{X_{gait}}$, $\Phi^{\prime}_{crab} = \Phi^{\prime}_{gait}$ given by Eq. \hyperref[eq:eq8]{(8)}:
\begin{equation}\label{eq:eq8}
  \begin{aligned}
    \bm{\Phi^{\prime}_{walk}} &=
    \begin{bmatrix}
      0 & \frac{3}{2}\pi & \pi & \frac{\pi}{2} \\
      -\frac{3}{2}\pi & 0 & -\frac{\pi}{2} & -\pi \\
      -\pi & \frac{\pi}{2} & 0 & -\frac{\pi}{2} \\
      -\frac{\pi}{2} & \pi & \frac{\pi}{2} & 0
    \end{bmatrix},
    \bm{\Phi^{\prime}_{turning}} =
    \begin{bmatrix}
      0 & -\frac{3}{2}\pi & -\frac{\pi}{2} & -\pi \\
      \frac{3}{2}\pi & 0 & \pi & \frac{\pi}{2} \\
      \frac{\pi}{2} & -\pi & 0 & -\frac{\pi}{2} \\
      \pi & -\frac{\pi}{2} & \frac{\pi}{2} & 0
    \end{bmatrix}, \\
    \bm{R_{crab}} &=
    \begin{bmatrix}
      0 & \frac{\pi}{18} & \frac{\pi}{9} & \frac{\pi}{9} \\
      0 & \frac{\pi}{18} & \frac{\pi}{9} & \frac{\pi}{9} \\
      0 & \frac{\pi}{18} & \frac{\pi}{9} & \frac{\pi}{9} \\
      0 & \frac{\pi}{18} & \frac{\pi}{9} & \frac{\pi}{9}
    \end{bmatrix}, \quad \quad \quad
    \bm{\Phi_{crab}} =
    \begin{bmatrix}
      0 & 0 & -\pi & 0 \\
      0 & 0 & -\pi & 0 \\
      0 & 0 & \pi & 0 \\
      0 & 0 & \pi & 0
    \end{bmatrix}.
  \end{aligned}
\end{equation}
These swimming and gait locomotion parameters are generated by the behavior transition system.
\subsection{Behavior transition system}
This approach enables multifunctional locomotion control as a nonlinear emergence based on sensor modalities, rather than through on/off switching via control commands or linear compensation. The swimming and gait behaviors are realized in three states:
\begin{enumerate}
  \item Robot state estimation using external sensors.
  \item Determines the target of behavior transition from estimated values.
  \item Executing motions based on the robot's state.
\end{enumerate}
The robot's state is estimated as the current water depth category based on the distance $D$ to the seafloor surface during initial submersion and the current depth $h$. For depth estimation, three subspaces, the surface, mid, and deep-layer, are evenly divided. This is estimated using the trapezoidal, triangular, and trapezoidal membership functions $\mu_i(h)$. The inference results for each membership function are approximated using the centroid method, as shown in Eq. \hyperref[eq:eq9]{(9)}:

\begin{equation}\label{eq:eq9}
  \begin{aligned}
    \hat{h} &= \frac{\sum_{i=0}^{N-1} \mu_i(h) \cdot c_i}{\sum_{i=0}^{N-1} \mu_i(h)}, \quad
    \hat{z} = \frac{\hat{h}}{D}.
  \end{aligned}
\end{equation}

Perception of the external environment utilizes point cloud data from an RGB-D camera mounted on the head. To determine the terrain of the external environment, DBSCAN is used. Clustering and size calculation is implemented for PCA labeled for floor, ceiling, wall, and object. Environments are estimated by adding an angle constraint on normal vectors to the DBSCAN distance threshold $N_{Eps}$. The normal vectors of each point cloud are denoted by $\mathbf{n}_i$ and $\mathbf{n}_j$, and $\theta_{threshold}$ is the angular threshold for the normal vectors. Visual information $\hat{s}$ is determined using the sigmoid function with the cluster size $c$ based on the detection result. The parameter $c_{mid}$ is defined as $c_{mid} = (c_{min} + c_{max})/2$, using the upper and lower bounds of the cluster size. The visual information $\hat{s}$ is defined as the wall cluster detection rate in Eq. \hyperref[eq:eq10]{(10)}:

\begin{equation}\label{eq:eq10}
  \begin{aligned}
    x_q &\leq N_{Eps} \quad AND \quad \theta_{normal}(\mathbf{n}_i, \mathbf{n}_j) \leq \theta_{threshold}, \\
    \hat{s} &= \frac{1}{1 + e^{-k(c-c_{mid})}}.
\end{aligned}
\end{equation}

The motion generation target is determined based on depth estimation results: swimming motion is selected for surface layers, and gait motion for deep layers. The parameter of $\Theta$ is the vector of installed in the target value for each generate the motion, $\Theta = [\bm{R}, \bm{X}, \bm{\Phi}, \bm{\Phi^{\prime}}]^T$, given by Eq. \hyperref[eq:eq11]{(11)}:

\begin{equation}\label{eq:eq11}
  \begin{aligned}
    \mathbf{\psi_d} &= P(\Theta_{swim}, \Theta_{gait}, \hat{z}) \\
    &= (\Theta_{gait} - \Theta_{swim}) \hat{z} + \Theta_{swim}.
  \end{aligned}
\end{equation}

Multiple behaviors and sensor modalities can be realized by providing a potential function for the desired behavior. In this proposal, we designed a potential function using three multifunction and two sensor inputs as Eq. \hyperref[eq:eq12]{(12)}. The visual information input $\hat{s}$ is used to generate different gait patterns via the weight function $w(\hat{s})$.

\begin{equation}\label{eq:eq12}
  \begin{aligned}
    \mathbf{\psi_d} &= P(\Theta_{swim}, \Theta_{gait}, \Theta_{turning}, \hat{z}, \hat{s}), \\
    &= -sin((\hat{z} - z_t)\frac{\pi}{2})\Theta_{swim} + cos((\hat{z} - z_t)\frac{\pi}{2})\Theta_{gait}, \\
    \Theta_{gait} &= w(\hat{s})\Theta_{turning} + (1-w(\hat{s}))\Theta_{trot}, \\
    w(\hat{s}) &= \frac{1+\cos(\pi \hat{s})}{2}.
  \end{aligned}
\end{equation}

The design of potential functions can easily generate multiple dynamics and can be considered in gradient systems.
\begin{equation}\label{eq:eq13}
  \begin{aligned}
    \frac{d\Theta_{swim}}{dt} &= -\frac{\partial P}{\partial \Theta_{swim}} = sin((\hat{z} - z_t)\frac{\pi}{2}), \\
    \frac{d\Theta_{gait}}{dt} &= -\frac{\partial P}{\partial \Theta_{gait}} = -cos((\hat{z} - z_t)\frac{\pi}{2}). \\
  \end{aligned}
\end{equation}

A transition to target behaviors is performed using a potential function that does not possess multiple minima and permits arbitrary changes in gradient given by Eq. \hyperref[eq:eq14]{(14)}. This ensures that undesirable movement patterns are not generated during transitions between target behaviors \citep{yuasa1990coordination, odashima2002hierarchical}. A Gaussian error function embedded in torus space is used as the potential function.
\begin{equation}\label{eq:eq14}
  \begin{aligned}
    W(\psi_i) &= -g \exp[-T(1 - \cos(\mathbf{\psi_{i}} - \mathbf{\psi_{d,i}}))], \\
    \dot{W(\psi_i)} &= T \sin(\mathbf{\psi_{i}} - \mathbf{\psi_{d,i}})W(\psi_i).
  \end{aligned}
\end{equation}

The parameter $T$ represents the gradient of the potential function at the valley, and $g$ is the gradient of the potential function. The parameters $\mathbf{\psi_{i}}$ and $\mathbf{\psi_{d,i}}$ represent the current value and target value of each CPG parameter, respectively.
\begin{equation}\label{eq:eq15}
  \begin{aligned}
\mathbf{\psi_{t+1,d}} &= \mathbf{\psi_{t,d}} - \alpha \dot{W(\psi_i)}.
  \end{aligned}
\end{equation}
This gradient value is used to update the CPG parameters. Parameter $\alpha$ is the update coefficient, set to 0.95.

\begin{figure}[t]
  \centering
  \resizebox*{10cm}{!}{\includegraphics{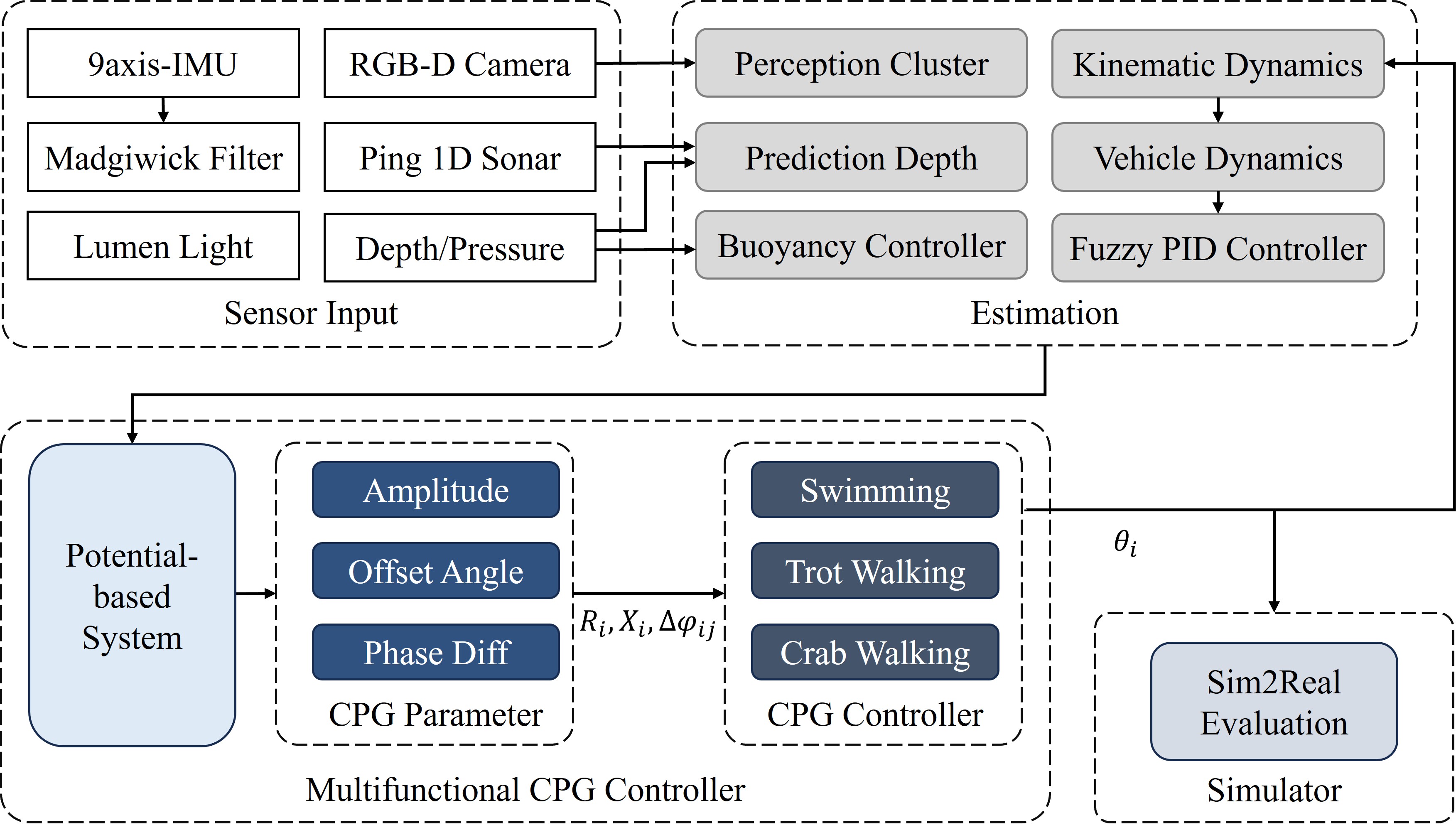}}\hspace{5pt}
  \caption{Software Architecture for the Entire System} \label{fig:Fig4}
\end{figure}

The overall software architecture of the multifunctional locomotion controller is shown in Fig. \hyperref[fig:Fig4]{4}. The robot detects and estimates depth and wall clusters using multiple sensor modalities. The estimated results are input into the potential-based motion generator, which uses multifunction to nonlinearly emerge behaviors according to the state. These software architectures are implemented using Robot Operating System (ROS) 2, and their basic locomotion performance can be verified via simulator.

\section{Experiment and Result}
\label{sec:Sec4}
\begin{figure}[t]
  \centering
  \resizebox*{8cm}{!}{\includegraphics{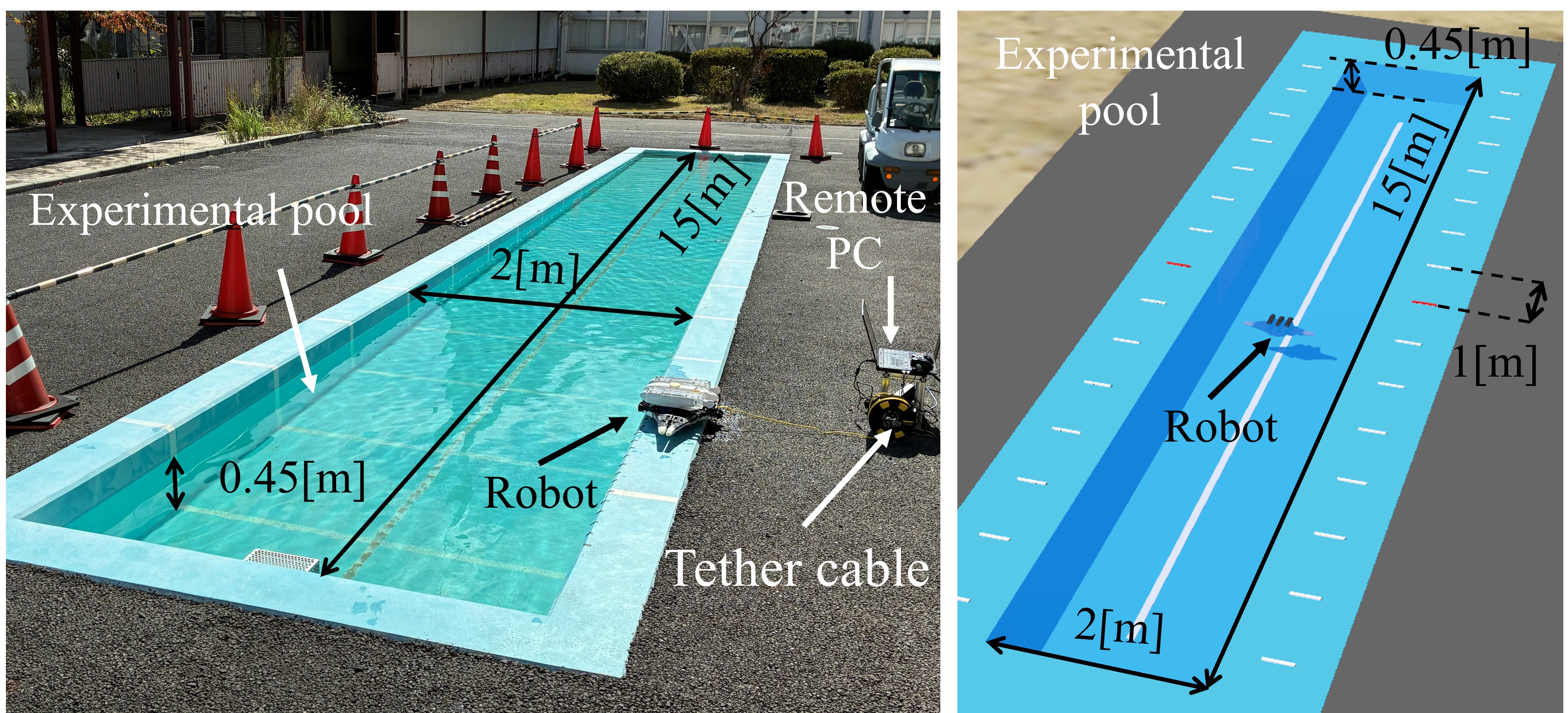}}\hspace{5pt}
  \caption{Experimental pool in real and simulation environments.} \label{fig:Fig5}
\end{figure}
\subsection{Swimming ang gait locomotion}
\begin{figure}[t]
  \centering
  \subfloat[]{
  \resizebox*{7cm}{!}{\includegraphics{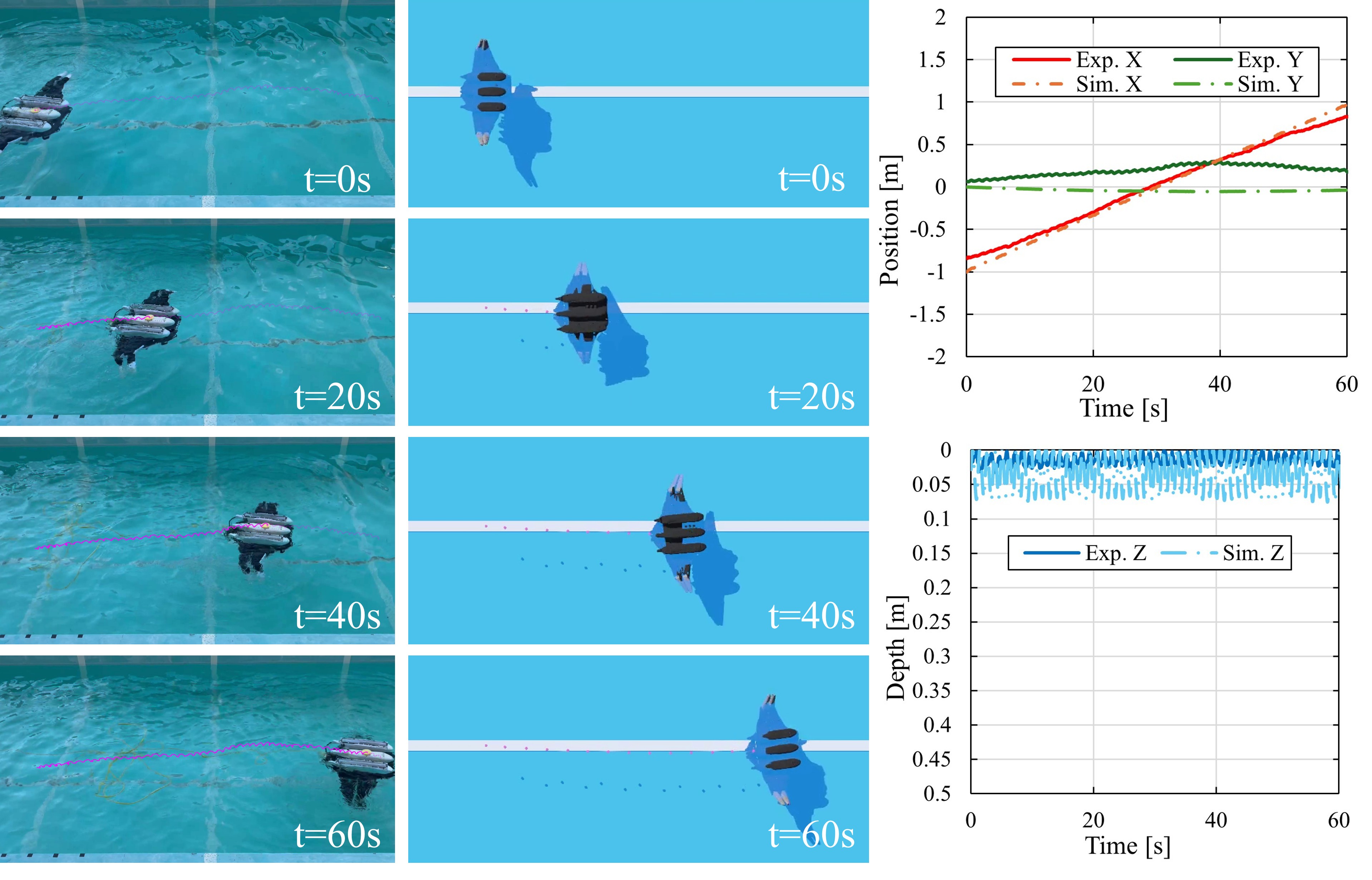}}}\hspace{5pt}
  \subfloat[]{
  \resizebox*{7cm}{!}{\includegraphics{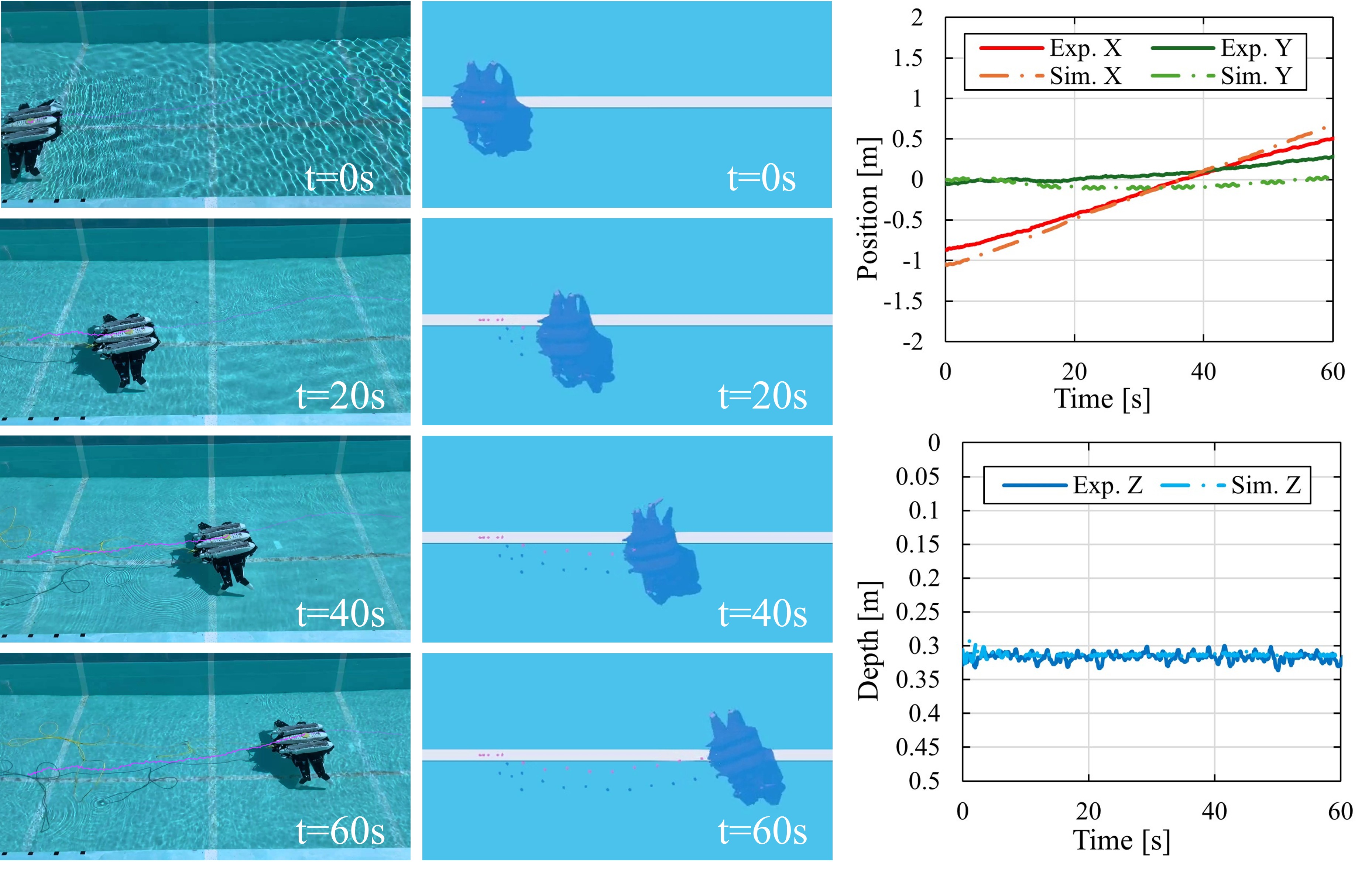}}}\hspace{5pt}
  \subfloat[]{
  \resizebox*{7cm}{!}{\includegraphics{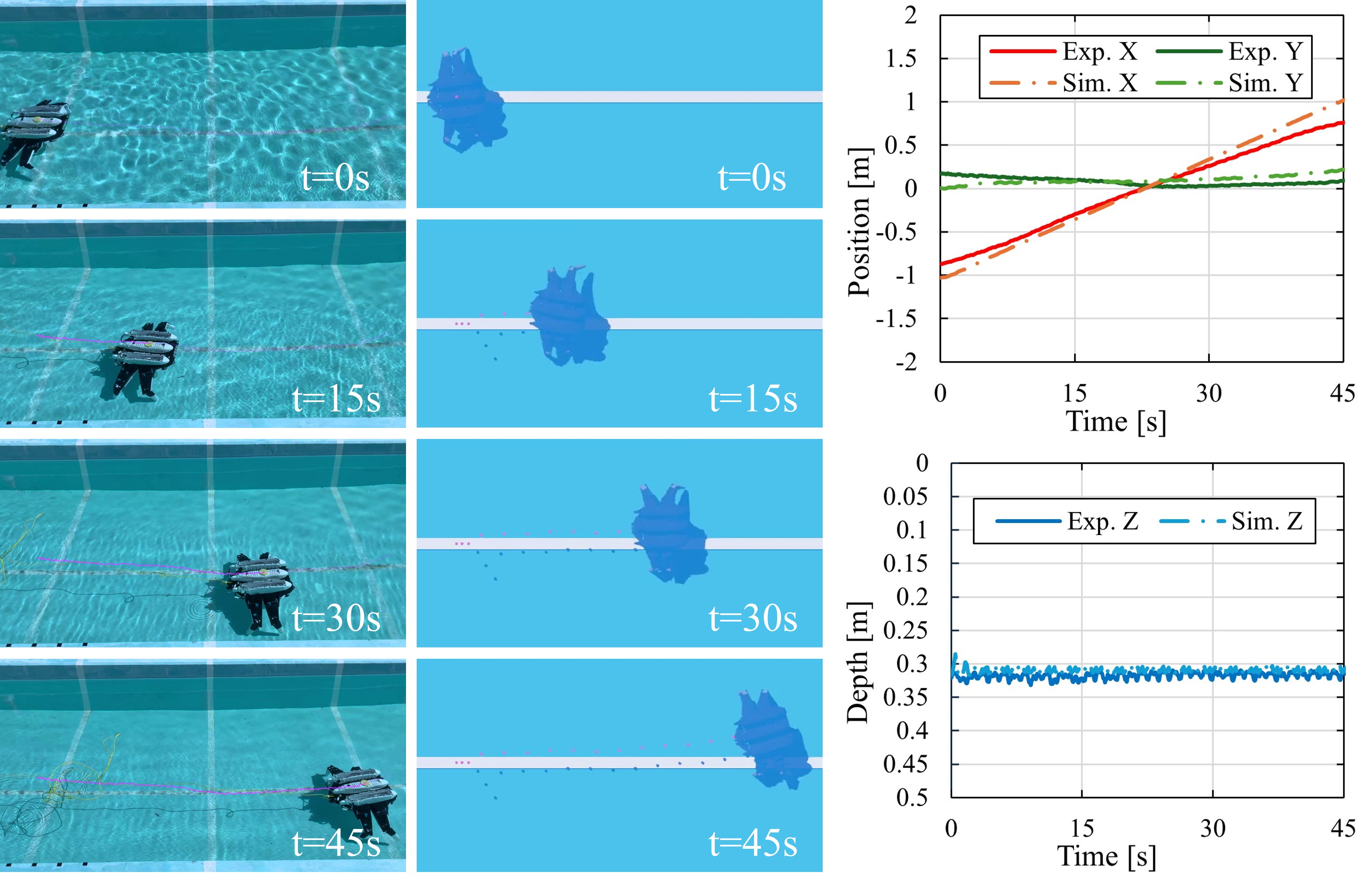}}}\hspace{5pt}
  \caption{Comparison of experimental and simulated motions in locomotion (a) swimming, (b) trot gait, (c) walk gait.} \label{fig:Fig6}
\end{figure}

\begin{figure}[t]
  \centering
  \resizebox*{10cm}{!}{\includegraphics{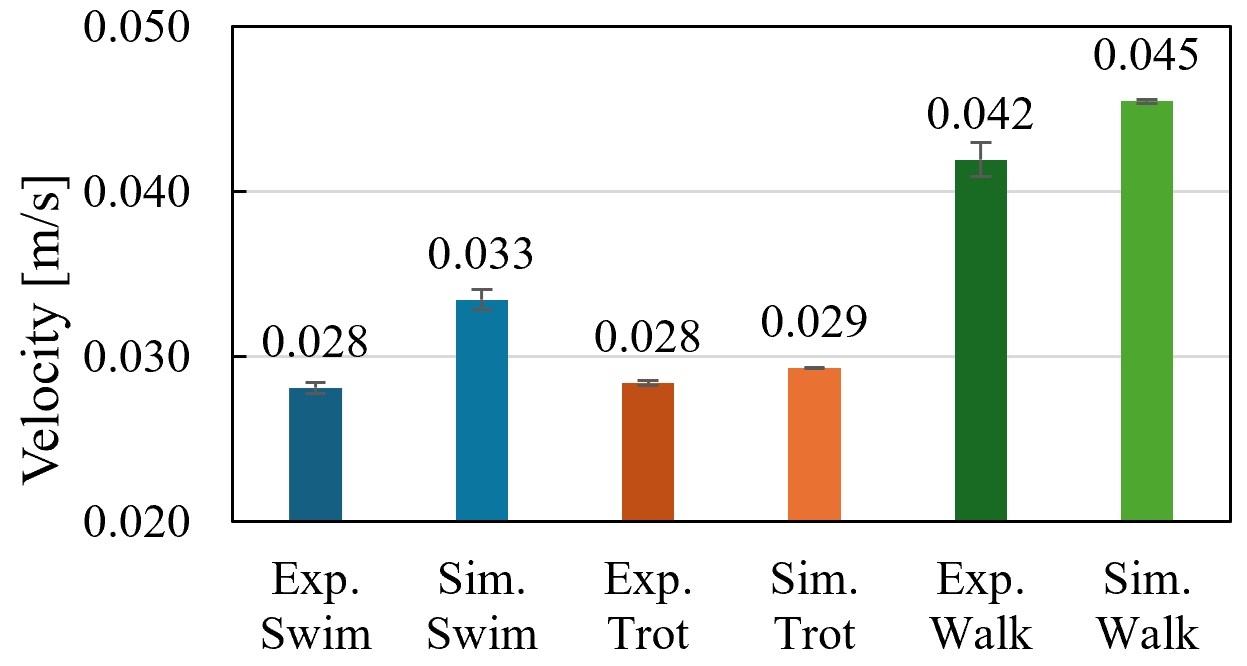}}\hspace{5pt}
  \caption{Comparision of locomotion velcity in experiments and simulations.} \label{fig:Fig7}
\end{figure}

\begin{figure}[t]
  \centering
  \resizebox*{10cm}{!}{\includegraphics{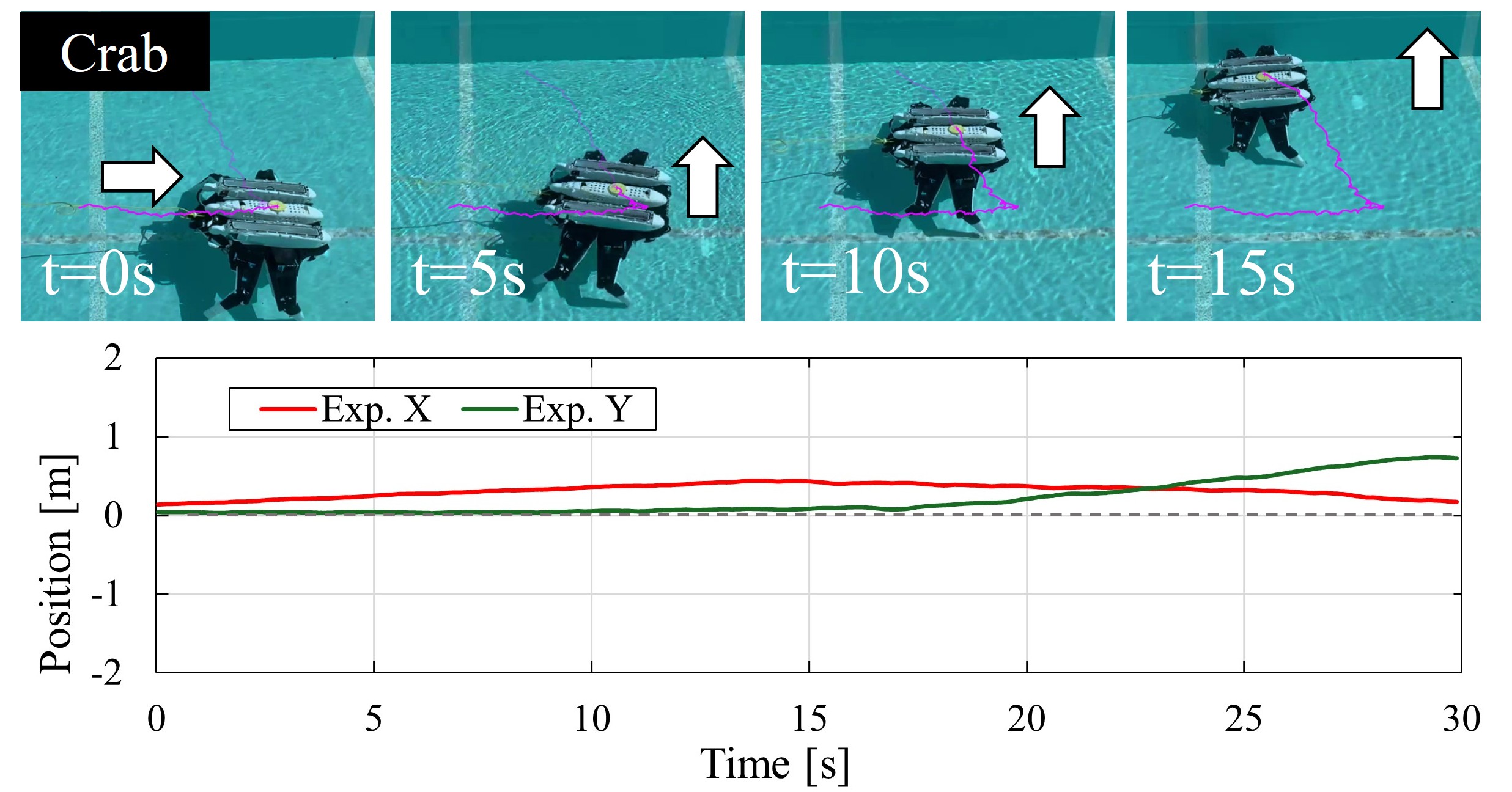}}\hspace{5pt}
  \caption{Crab walk ability of the optional and time variation of the position.} \label{fig:Fig8}
\end{figure}

Experiments are conducted in both the real and simulated environments in Fig. \hyperref[fig:Fig5]{5} to evaluate the performance of each locomotion method. The experimental environment is used in an outdoor pool. The dimensions of the experimental pool are 15 [m] in length, 2 [m] in width, and 0.5 [m] in depth. The experimental method evaluates performance for fundamental locomotion, based on swimming and gait (trot, walk, and crab gait). The simulation environment uses Webots, used for BUR evaluation \citep{zhangDynamicTargetTracking2024a, struebig2020design}. The results from the simulation environment and the measurements from the physical machine minimize the gap by looping Sim2Real \citep{tasadaDolphin}. In the simulation environment, the density of water and viscous torque are set to 1000 $kg/m^3$ and 5, respectively. Linear and angular damping were optimized through iterative Sim2Real runs and set to 0.8, respectively. Swimming and gait locomotion are measured two times each using actual measurements and simulations. Figure \hyperref[fig:Fig6]{6} shows representative snapshots of multimodal locomotion, while Fig. \hyperref[fig:Fig7]{7} shows the velocities of each locomotion. Furthermore, Fig. \hyperref[fig:Fig8]{8} shows the representative measurement results of crab gait capable of emerging from the same multi-jointed structure.

\begin{figure}[t]
  \centering
  \resizebox*{10cm}{!}{\includegraphics{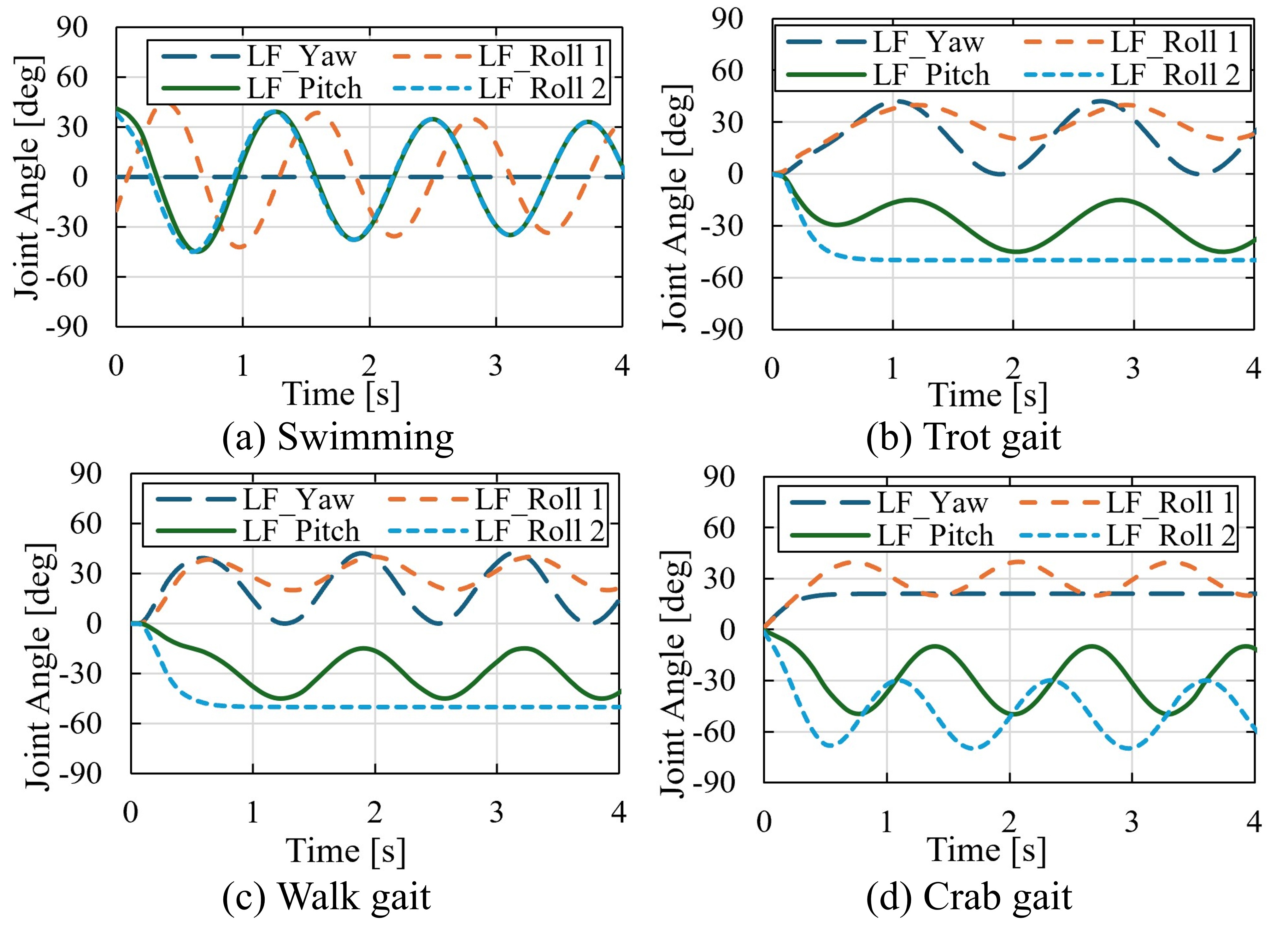}}\hspace{5pt}
  \caption{Comparison of joint angle data for the left front leg-fin, (a) swimming, (b) trot gait, (c) walk gait, (d) crab gait.} \label{fig:Fig9}
\end{figure}

Experimental results confirmed the generation of swimming, trot, walk, and crab gait motions from the same multi-jointed structure. This shows that various movements can be generated depending on the joint angles of the multi-joint structure. From the position variation data, forward movement was observed in both swimming and gait locomotion. The measured values for position and velocity were approximately equal to the results obtained in the simulation environment. The walk gait achieved the fastest speed among all locomotion. The pectoral fins are thicker and smaller than those of previous proposed manta-ray robots \citep{asada2024development}, so the swimming speed is lower. The relationship would be considered a trade-off with gait speed. One leg-fin has four joints (yaw, roll, pitch, and roll), which could be considered a structure suitable for gait locomotion. Figure \hyperref[fig:Fig9]{9}  hows a comparison of the joint angle data for the left front leg-fin. From the joint angle data, it can be confirmed that the amplitude, offset, and phase difference defined in Eq. \hyperref[eq:eq6]{(6)}--\hyperref[fig:Fig8]{(8)} are achieved. Two-layer CPG control and parameter design can be considered as key factors in generating different functions.

\begin{figure*}[t]
  \centering
  \subfloat[]{%
    \includegraphics[width=0.8\textwidth]{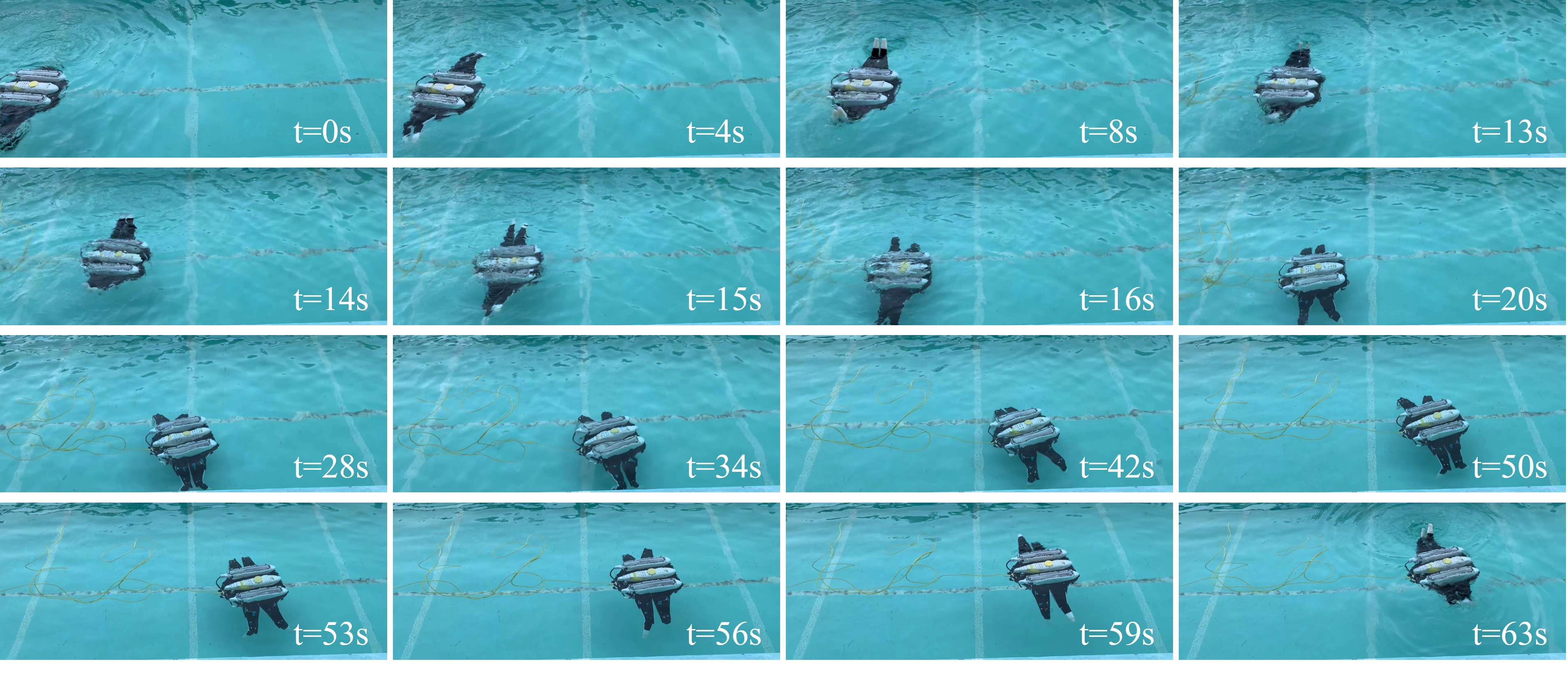}
  }\hspace{5pt}
  \subfloat[]{%
    \includegraphics[width=0.8\textwidth]{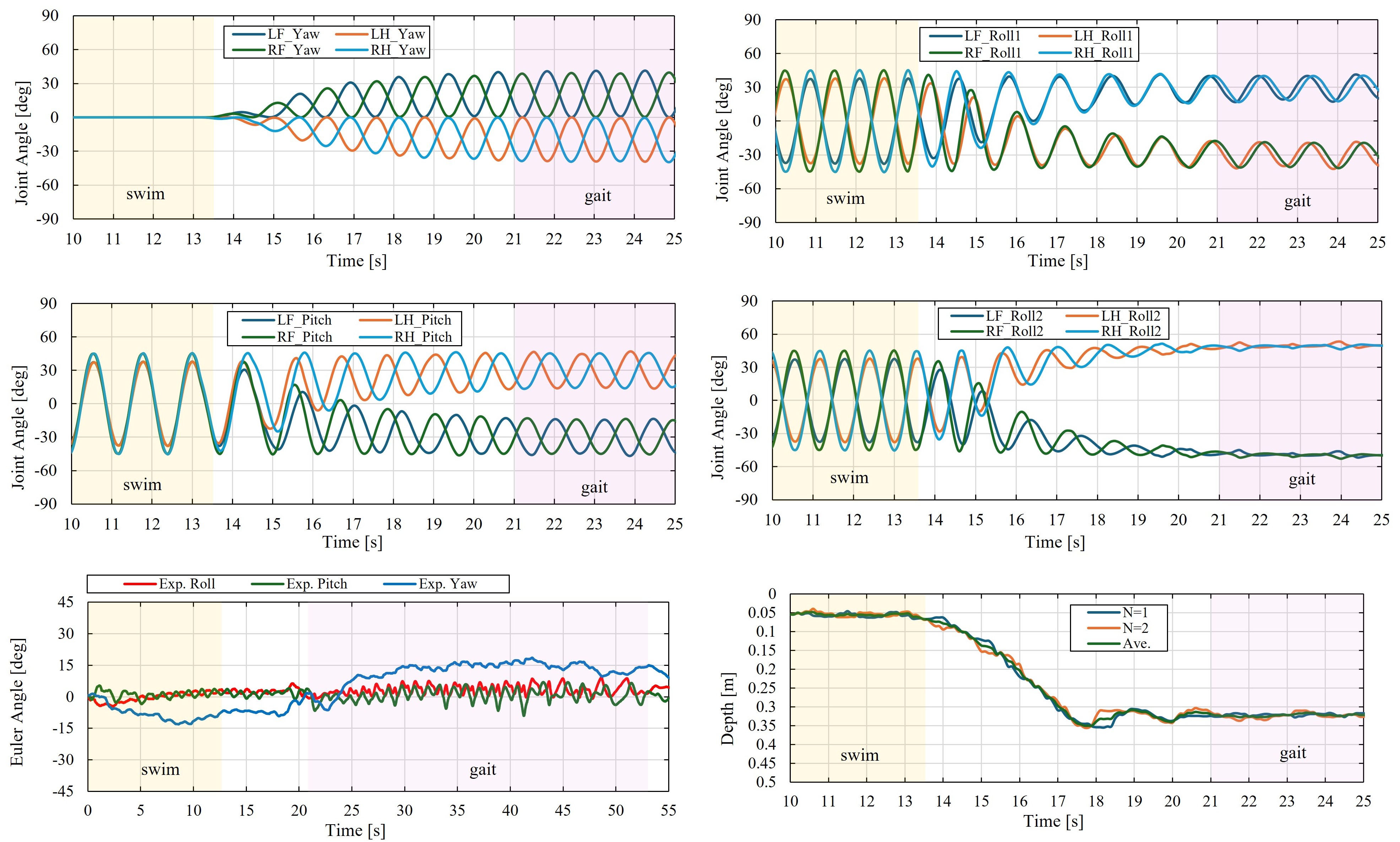}
  }
  \caption{Multifunctional locomotion of two motion generation. (a) snapshot of the motion, (b) output signals of each joint, euler, and depth data.}
  \label{fig:Fig10}
\end{figure*}

\subsection{Two behavior transition of swimming and gait}
Validate the transition of swimming and trot gait as the potential-based control algorithm. The transition of joint angles from swimming to trot gait is evaluated during these experiments. Potential functions are used to verify that joint angle is not transitioned to unexpected values. Figure \hyperref[fig:Fig10]{10} shows a representative snapshot of the two behavior transition and joint angles data. From Fig. \hyperref[fig:Fig10]{10}(a) a smooth transition between swimming and gait joint outputs can be observed from 13[s] to 21[s].  The transition from gait locomotion to swimming was also observed between 50[s] and 63[s] in Fig. \hyperref[fig:Fig10]{10}(b). This is because the function of the behavior transition target is used as a preprocessing step for the potential function of the transition of behavior. In each mode of Euler angles, it was confirmed that swimming exhibited smaller variation in roll and pitch angles. In swimming, the robot's attitude remains stable at the water surface due to the positive buoyancy provided by BVBS. In gait locomotion, negative buoyancy is generated, and the kicking of the robot during the stepping phase is affected. The depth and time variation causes minute displacement along the z-axis. This also affects the small changes in Roll2 joint angles, which are observed as gait errors.

The smooth behavioral transitions between swimming and gait locomotion were achieved as designed by the potential functions in Eq. \hyperref[eq:eq14]{(14)} and \hyperref[eq:eq15]{(15)}. Embedding the Gaussian error function into torus space would complement and emerge the intermediate locomotion between swimming and trot gait between 13 [s] and 21 [s]. With a single sensor modality, the decision of the behavior transition target is dominated by the depth prediction $\hat{z}$. This approach is suitable for scenarios when behavior changes in response to a single sensor or control input.
\begin{figure*}[t]
  \centering
  \subfloat[]{%
    \includegraphics[width=0.49\textwidth]{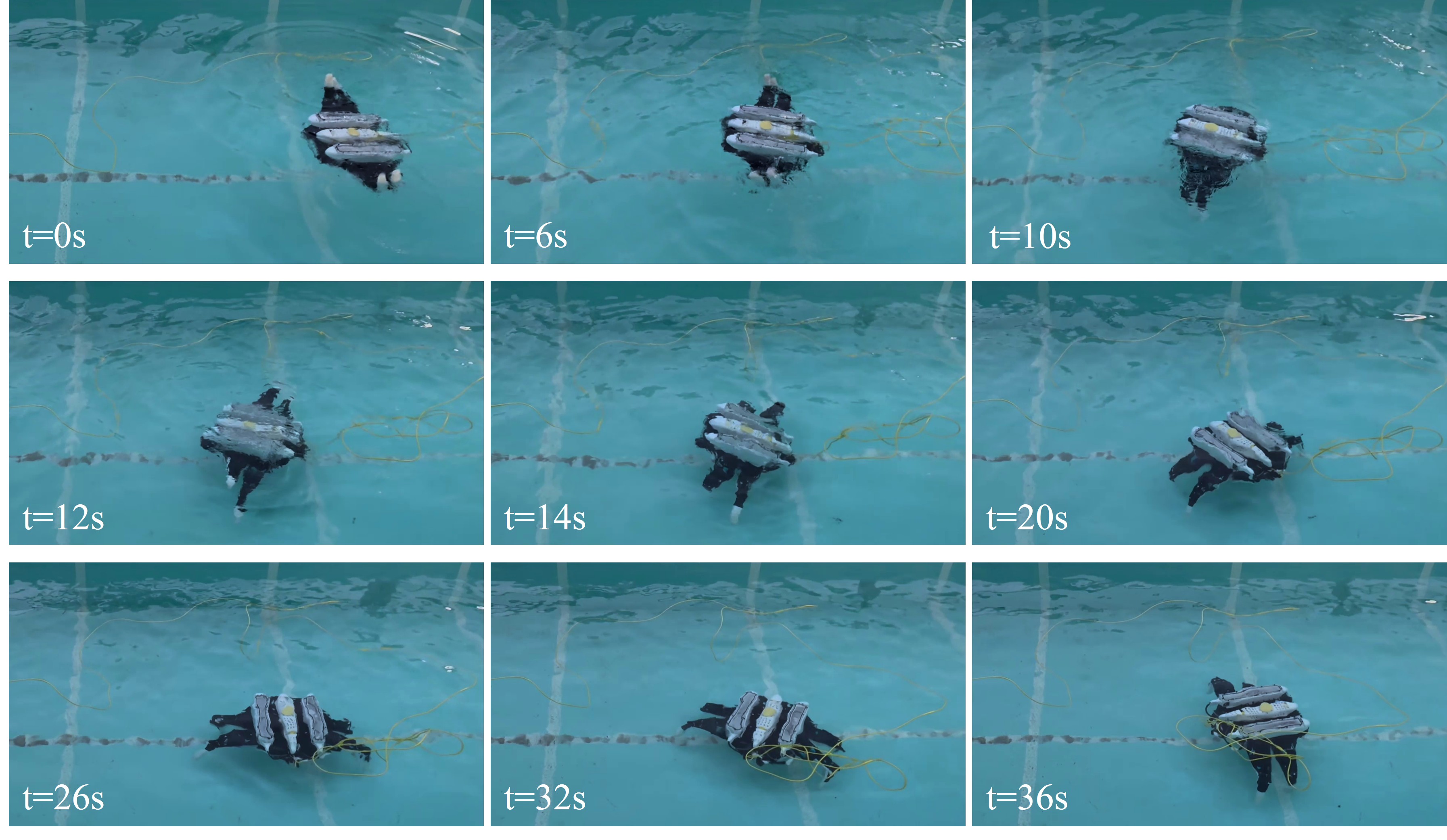}
  }\hspace{5pt}
  \subfloat[]{%
    \includegraphics[width=0.46\textwidth]{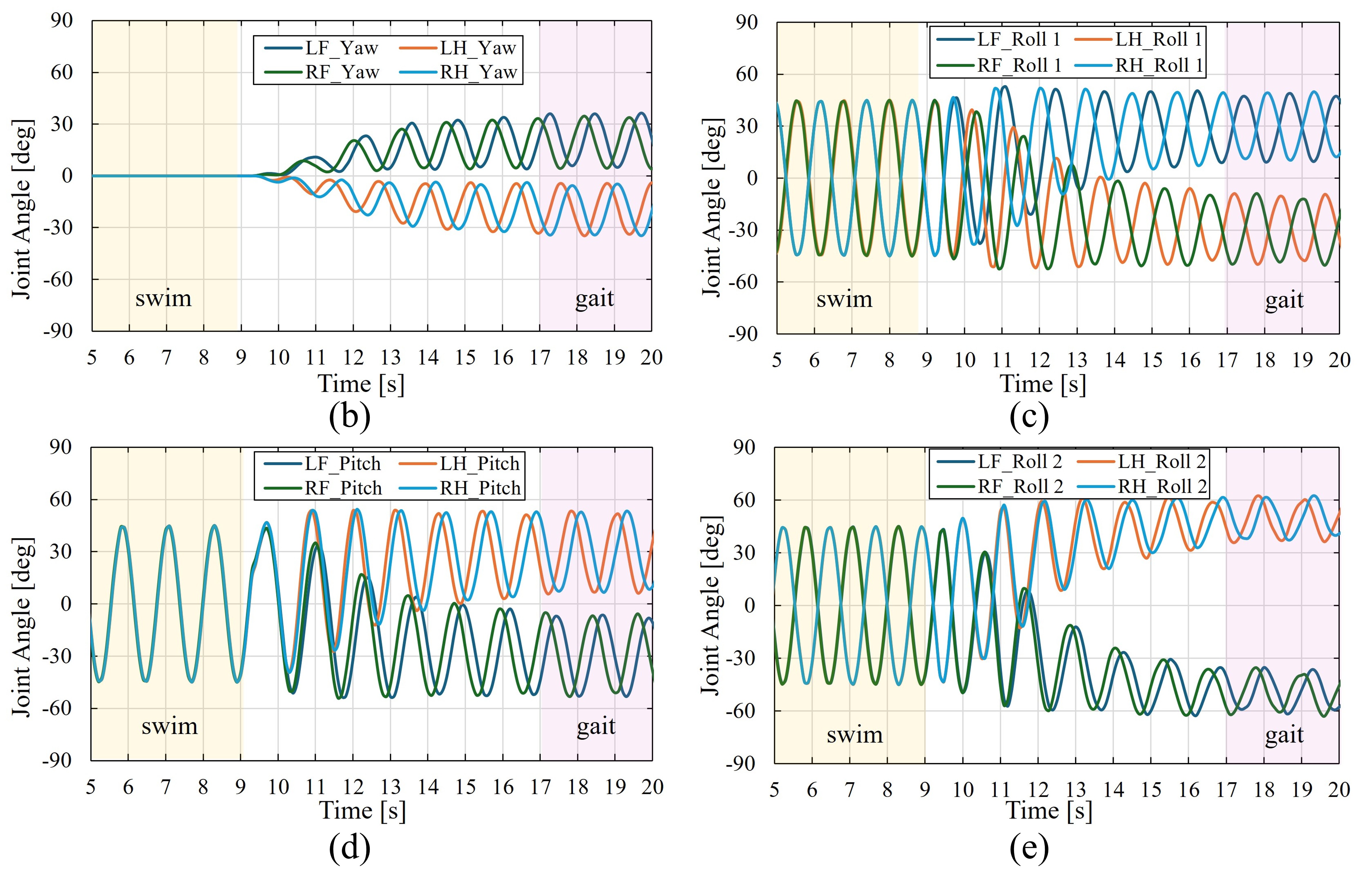}
  }
  \caption{Multifunctional locomotion of multimodal sensory feedback.  (a) snapshot of the locomotion, (b) output signals of each joint.}
  \label{fig:Fig11}
\end{figure*}

\subsection{Multifunctional generation of three behaviors}
A multifunctional transition system is validated, using three locomotion behaviors and two sensor inputs. The behaviors are defined as swimming, trot gait, and turning gait. The sensor modality uses predicted depth and visual information. Visual information is designed such that the parameters change when a walk cluster is detected. A potential function is also used for the behavior transition target. Thus, the target switches between swimming and gait locomotion modes based on depth prediction. In gait locomotion mode, transition between trot and turning gait is performed based on the cluster results. This verifies whether the target's behavior can be expressed. The experiment is conducted two times at the experimental pool. Representative snapshots and experimental data are shown in Fig. \hyperref[fig:Fig11]{11}. As shown in Fig. \hyperref[fig:Fig11]{11}(a), the transition from swimming to a trot gait was observed between 6[s] and 14[s]. Between 20[s] and 36[s], the transition from trot gait to turning gait was observed. From 9[s] to 17[s] in Fig. \hyperref[fig:Fig11]{11}(b), it was confirmed that the transition of each joint state is transitioned to the target joint angle. This detected the target cluster from the visual information, demonstrating that the behavior transitioned as designed. A representative example of detection results using clustering is shown in Fig. {12}.

As shown in Eq. \hyperref[eq:eq12]{(12)}, we verified that swimming and gait are transitioned by depth estimates, while trot and turning gait are driven by visual information. Define the sensor modalities corresponding to the target behaviors in advance, enabling the transition between multiple behaviors. Intermediate behaviors are supplemented by a potential function depending on how sensor modalities and behaviors are combined. However, as the number of modalities and behaviors increases, there are limitations that complicate the definition of the motion generation target.
\begin{figure}[t]
  \centering
  \resizebox*{8cm}{!}{\includegraphics{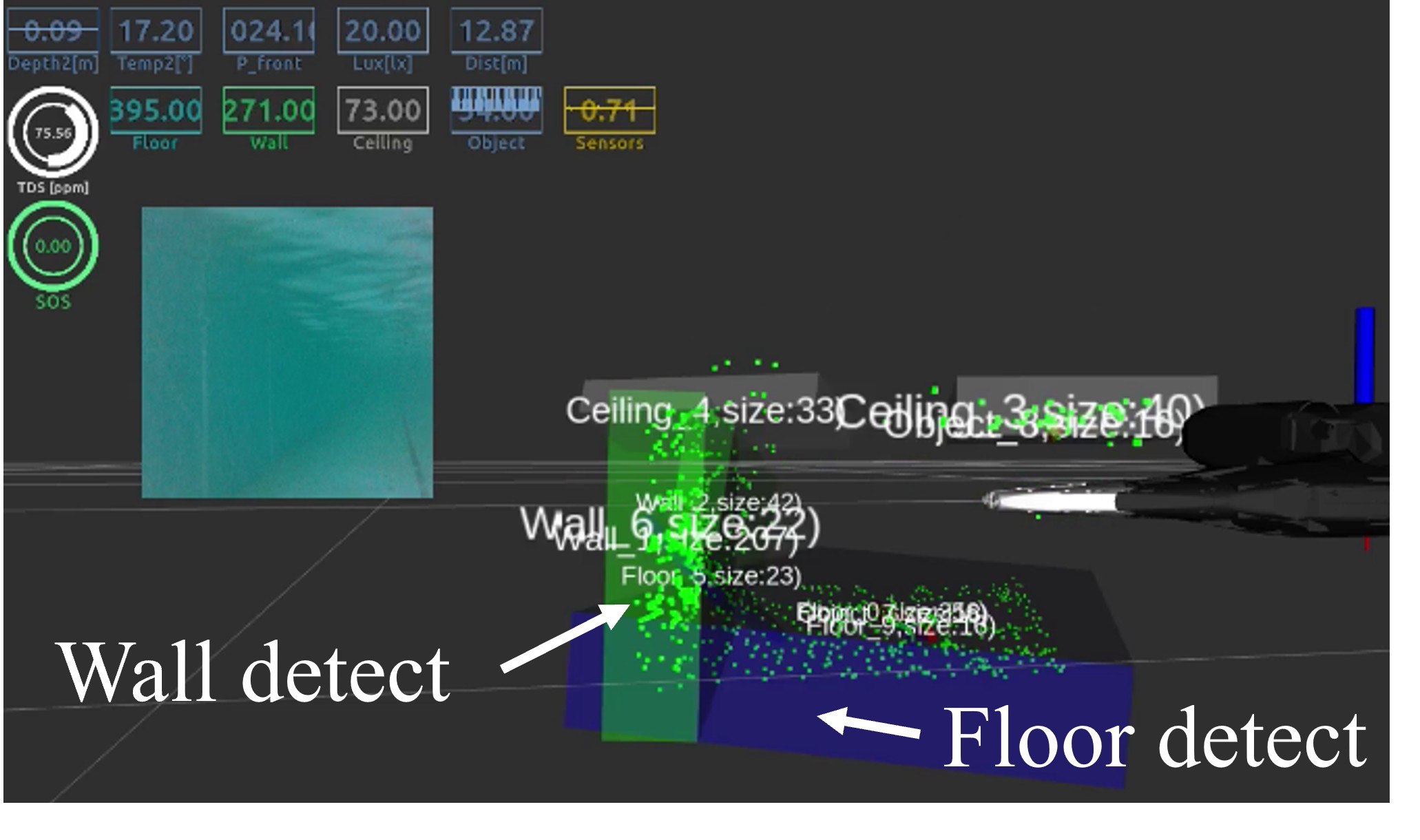}}\hspace{5pt}
  \caption{Clustering Detection Results for Walls and Floors in water.} \label{fig:Fig12}
\end{figure}

\subsection{Potential applications}
The multifunctional locomotion controller can realize multifunctionality from a simple multi-joint structure by utilizing sensor modalities Multi-joint mechanisms can achieve motions beyond biological capabilities by appropriately designing the amplitude, offset, and phase difference of the oscillator. The robot can perform various behaviors without depending on biomimicry or specialized mechanisms. Thus, this system provides extensibility by installing multi-joint for robots, enabling the control system to design multifunctional capabilities. The design of a variety of sensor modalities is limited by the differences in the sensors that can be mounted on a robot. Transition theory based on potential functions is independent of sensor modality, whereas it forms the core of transition to target behavior. Its multi-joint structure enables expansion to many biomimetics robots as multifunctional control. Table \hyperref[table:tbl3]{3} clearly depicts the potential of the proposed method and robot compared to BURs capable of multifunctional expression. Conventional multifunctional BURs include not only MPF type with symmetrical pectoral fins, but also BCF type utilizing the body and flipper. This indicates that the corresponding functions are required to achieve the desired multifunctionality. In other words, the ability to achieve multifunctionality from a single MPF structure indicates high scalability, not dependent on the designer's discretion. Focusing on mechanisms and material properties, conventional robots are equipped with their own unique mechanisms. In multifunctional control, few clearly specify multifunctional control algorithms, with most relying on manual operation or switching. Our proposed robot and approach clearly show a significant advantage in terms of emergent capabilities from multi-joint systems, rather than possessing multiple functions in advance.

\begin{table*}[t]
\centering
\caption{Comparison of the multimodal BURs.}
\scalebox{0.56}{
\begin{tabular}{cccccccccccc}
\hline
Name  & Mass & Buoy. & \multicolumn{4}{c}{Propulsion Type $\&$ DOF} & Mode & \multicolumn{3}{c}{Multifunction} & Mechanical \\
\cline{4-7}\cline{9-11}
  & & control & Total & MPF & BCF & Flipper & control & Swimming & gait & Etc. & and Material \\
\hline
Ours & 8.4 & \checkmark & 16 & $4\times 4$ & - & - & \makecell{\checkmark \\ (Potential)} & \checkmark & \makecell{\checkmark \\ (Water)} & \makecell{Crab Walk \\ (Water)} & Multi-joint \\
\hline
BoxyBot \citep{crespi2008controlling} & - & - & 3 & - & 1 & $1\times 2$ & \makecell{\checkmark \\ (Manual)} & \checkmark & - & \makecell{Crawling \\ (Land)} & Multi-joint \\
\hline
\makecell{Salamandra \\ Robotica II \citep{crespi2013salamandra}} & 2.5 kg & - & 12 & $1\times 4$ & 8 & - & \makecell{\checkmark \\ (Manual)} & \checkmark & \makecell{\checkmark \\ (Land)} & - & \makecell{Individual \\ (MPF, BCF)} \\
\hline
CR200 \citep{yoo2016design} & 682 kg & - & 24 & $4\times 6$ & - & - & - & \checkmark & \makecell{\checkmark \\ (Water)} & - & Multi-joint \\
\hline
ART \citep{baines2022multi}& 9 kg & - & 12 & $3\times 4$ & - & - & - & \checkmark & \makecell{\checkmark \\ (Land)} & \makecell{Crawling \\ (Land)} & \makecell{Adaptive \\ Morphogenesis} \\
\hline
PEAR \citep{kim2021underwater} & 7.15 kg & \checkmark & 6 & $1\times 6$ & - & - & \makecell{\checkmark \\ (Manual)} & \checkmark & \makecell{\checkmark \\ (Water)} & - & \makecell{Hinged Multi-Modal \\ Paddle} \\
\hline
HERO-BLUE \citep{kim2024development} & 11.32 kg & \checkmark & 9 & $1\times 4$ & 1 & $2\times 2$ & \makecell{\checkmark \\ (Manual)} & \checkmark & \makecell{\checkmark \\ (Water)} & \makecell{Crawling \\ (Water)} & \makecell{Passive Locking \\ Mechanism} \\
\hline
BRIM \citep{song2023development} & 18.6 kg & - & 8 & $2\times 2$ & - & $1\times 4$ & \makecell{\checkmark \\ (Manual)} & \checkmark & \makecell{\checkmark \\ (Surface)} & \makecell{Pushing \\ (Obstacle)} & \makecell{Hinged fins} \\
\hline
NCUUV \citep{yan2020research} & 42.5 kg & \checkmark & 21 & $3\times 6$ & 1 & $1\times 2$ & \makecell{\checkmark \\ (Switching)} & \checkmark & \makecell{\checkmark \\ (Water)} & - & \makecell{Individual \\ (MPF, BCF, Flipper)} \\
\hline 	\label{table:tbl3}
\end{tabular}
}
\end{table*}

\section{CONCLUSIONS}
\label{sec:Sec5}
In this study, multifunctional locomotion control for multi-jointed BURs were verified. The design was implemented as a multi-joint system for organisms that traditionally have no multifunctional capabilities. The different behaviors of swimming, trot gait, walk gait, and crab gait were generated by setting parameters using a two-layer CPG controller. As a control system, we experimented with multifunctional generation and behavioral transitions using multiple sensor modalities. The potential function is used to implement the mechanism for controlling multifunctional expression. It was demon-
strated possible to transition to up to three behaviors using estimated depth and visual information results as sensor modality inputs. This indicates that multi-joint design and multifunctional control design are expandable to generate multiple functions. The
use of multi-jointed and multifunctional controllers in BURs is expected to expand their applications in underwater exploration. In future work, a control system capable of generating multiple behaviors for multiple sensor modalities will be developed. Multifunctional behavior decisions based on the robot's posture, flow velocity, and power consumption would achieve emergence suitable for the external environment.

\section*{Disclosure statement}
No potential conflict of interest was reported by the author(s).

\section*{Funding}
No funding was received for conducting this study.

\section*{Notes on contributor(s)}
\begin{minipage}{\textwidth}
\raggedright
\textbf{Takumi Asada} received the B.E. and M.E. degree in electronics and information engineering from Aichi Institute of Technology (AIT), in 2021 and 2023. From 2023 to 2025, he worked at DENSO Corporation. He is currently a Ph.D. student in Utsunomiya University. He is receiving JSPS's DC2 Fellowship. His research interests include biomimetic underwater robots (BURs), nonlinear oscillators, and mechanical structure design. He is a member of JSME.\\[1em]

\noindent\textbf{Hideo Furuhashi} (Member, IEEE) received the Ph.D. degree from Nagoya University. He is now working with the Department of Electronics and Electrical Engineering, AIT. His research interests include the measurement and applications of light, ultrasonic waves, robotics and virtual reality (VR). \\[1em]

\noindent\textbf{Kenta Tabata} received the B.S., M.S., and Ph.D. degrees from Kanazawa University, in 2019, 2021, and 2023, respectively. Since 2023, he has been an Assistant Professor with Utsunomiya University. He is a member of JSME. \\[1em]

\noindent\textbf{Renato Miyagusuku} (Member, IEEE) received the B.S. degree in mechatronics from the Department of Mechanical Engineering, National University of Engineering, Peru, in 2011, and the M.S. and Ph.D. degrees from the Department of Precision Engineering, The University of Tokyo, Japan, in 2015 and 2018, respectively. Since 2019, he has been an Assistant Professor with Utsunomiya University and a Visiting Researcher with The University of Tokyo, and later has been an Associate Professor, since 2025. His research interests include machine learning applied to robotics, sensor fusion, and robotic applications to agriculture. He is a member of JSME. \\[1em]

\noindent\textbf{Koichi Ozaki} received the Ph.D. degree from the Graduate School, Division of Science and Engineering, Saitama University, in 1996. Since 2011, he has been a Professor with Utsunomiya University. In 2018, he established and headed the Robotics, Engineering and Agriculture-Technology Laboratory. His research interests include agricultural robots and mobile robotics. He is a member of JSME, RSJ, and JSPE.
\end{minipage}

\bibliographystyle{tfnlm}
\bibliography{bibtex}

@book{Fossen2021,
  author    = {Fossen, Thor I.},
  title     = {Handbook of Marine Craft Hydrodynamics and Motion Control},
  edition   = {2nd},
  year      = {2021},
  publisher = {John Wiley \& Sons},
  address   = {West Sussex, UK},
  isbn      = {978-1-119-57505-4},
}

@article{ev1989principles,
  title={Principles of naval architecture},
  author={Lewis},
  journal={SNAME},
  pages={365},
  year={1989}
}

@book{newman2018marine,
  title={Marine hydrodynamics},
  author={Newman, John Nicholas},
  year={2018},
  publisher={MIT press}
}

@article{go2019hydrodynamic,
  title={Hydrodynamic derivative determination based on CFD and motion simulation for a tow-fish},
  author={Go, Gwangsoo and Ahn, Hyung Taek},
  journal={Applied Ocean Research},
  volume={82},
  pages={191--209},
  year={2019},
  publisher={Elsevier}
}

@article{sfakiotakis2002review,
  title={Review of fish swimming modes for aquatic locomotion},
  author={Sfakiotakis, Michael and Lane, David M and Davies, J Bruce C},
  journal={IEEE Journal of oceanic engineering},
  volume={24},
  number={2},
  pages={237--252},
  year={2002},
  publisher={IEEE}
}

@article{crespi2008controlling,
  title={Controlling swimming and crawling in a fish robot using a central pattern generator},
  author={Crespi, Alessandro and Lachat, Daisy and Pasquier, Ariane and Ijspeert, Auke Jan},
  journal={Autonomous Robots},
  volume={25},
  number={1},
  pages={3--13},
  year={2008},
  publisher={Springer}
}

@article{crespi2013salamandra,
  title={Salamandra robotica II: an amphibious robot to study salamander-like swimming and walking gaits},
  author={Crespi, Alessandro and Karakasiliotis, Konstantinos and Guignard, Andre and Ijspeert, Auke Jan},
  journal={IEEE Transactions on Robotics},
  volume={29},
  number={2},
  pages={308--320},
  year={2013},
  publisher={IEEE}
}

@article{yasui2019decoding,
  title={Decoding the essential interplay between central and peripheral control in adaptive locomotion of amphibious centipedes},
  author={Yasui, Kotaro and Kano, Takeshi and Standen, Emily M and Aonuma, Hitoshi and Ijspeert, Auke J and Ishiguro, Akio},
  journal={Scientific reports},
  volume={9},
  number={1},
  pages={18288},
  year={2019},
  publisher={Nature Publishing Group UK London}
}

@article{yan2020research,
  title={Research on motion mode switching method based on CPG network reconstruction},
  author={Yan, Zheping and Yang, Haoyu and Zhang, Wei and Gong, Qingshuo and Lin, Fantai},
  journal={IEEE Access},
  volume={8},
  pages={224871--224883},
  year={2020},
  publisher={IEEE}
}

@article{kim2024development,
  title={Development of bioinspired multimodal underwater robot “hero-blue” for walking, swimming, and crawling},
  author={Kim, Taesik and Kim, Juhwan and Yu, Son-Cheol},
  journal={IEEE Transactions on Robotics},
  volume={40},
  pages={1421--1438},
  year={2024},
  publisher={IEEE}
}

@article{kim2021underwater,
  title={Underwater walking mechanism of underwater amphibious robot using hinged multi-modal paddle},
  author={Kim, Taesik and Song, Young-woon and Song, Seokyong and Yu, Son-Cheol},
  journal={International Journal of Control, Automation and Systems},
  volume={19},
  number={4},
  pages={1691--1702},
  year={2021},
  publisher={Springer}
}

@article{baines2022multi,
  title={Multi-environment robotic transitions through adaptive morphogenesis},
  author={Baines, Robert and Patiballa, Sree Kalyan and Booth, Joran and Ramirez, Luis and Sipple, Thomas and Garcia, Andonny and Fish, Frank and Kramer-Bottiglio, Rebecca},
  journal={Nature},
  volume={610},
  number={7931},
  pages={283--289},
  year={2022},
  publisher={Nature Publishing Group UK London}
}

@article{wang2014cpg,
  title={CPG-based sensory feedback control for bio-inspired multimodal swimming},
  author={Wang, Ming and Yu, Junzhi and Tan, Min},
  journal={International Journal of Advanced Robotic Systems},
  volume={11},
  number={10},
  pages={170},
  year={2014},
  publisher={SAGE Publications Sage UK: London, England}
}

@article{zhang2021design,
  title={Design and locomotion control of a dactylopteridae-inspired biomimetic underwater vehicle with hybrid propulsion},
  author={Zhang, Tiandong and Wang, Rui and Wang, Yu and Cheng, Long and Wang, Shuo and Tan, Min},
  journal={IEEE Transactions on Automation Science and Engineering},
  volume={19},
  number={3},
  pages={2054--2066},
  year={2021},
  publisher={IEEE}
}

@article{yoo2016design,
  title={Design of walking and swimming algorithms for a multi-legged underwater robot crabster CR200},
  author={Yoo, Seong-yeol and Shim, Hyungwon and Jun, Bong-Huan and Park, Jin-Yeong and Lee, Pan-Mook},
  journal={Marine Technology Society Journal},
  volume={50},
  number={5},
  pages={74--87},
  year={2016},
  publisher={Marine Technology Society}
}

@article{chen2022study,
  title={Study on the design and experimental research on a bionic crab robot with amphibious multi-modal movement},
  author={Chen, Xi and Li, Jiawei and Hu, Shihao and Han, Songjie and Liu, Kaixin and Pan, Biye and Wang, Jixin and Wang, Gang and Ma, Xinmeng},
  journal={Journal of Marine Science and Engineering},
  volume={10},
  number={12},
  pages={1804},
  year={2022},
  publisher={MDPI}
}

@article{wu2024underwater,
  title={An underwater biomimetic robot that can swim, bipedal walk and grasp},
  author={Wu, Qiuxuan and Pan, Liwei and Du, FuLin and Wu, ZhaoSheng and Chi, XiaoNi and Gao, FaRong and Wang, Jian and Zhilenkov, Anton A},
  journal={Journal of Bionic Engineering},
  volume={21},
  number={3},
  pages={1223--1237},
  year={2024},
  publisher={Springer}
}

@article{asada2024development,
  title={Development of a manta ray robot with underwater walking function},
  author={Asada, Takumi and Furuhashi, Hideo},
  journal={Ocean Engineering},
  volume={308},
  pages={118261},
  year={2024},
  publisher={Elsevier}
}

@article{luo2019parametric,
  title={Parametric geometric model and shape optimization of airfoils of a biomimetic manta ray underwater vehicle},
  author={Luo, Yang and Pan, Guang and Huang, Qiaogao and Shi, Yao and Lai, Hui},
  journal={Journal of Shanghai Jiaotong University (Science)},
  volume={24},
  number={3},
  pages={402--408},
  year={2019},
  publisher={Springer}
}

@article{xiang2025variable,
  title={A Variable Stiffness Fin for Manta Ray-Inspired Robots With Two Motion Modals},
  author={Xiang, Yang and Gu, Le and Ye, Kangjie and Zhang, Zhenwei and Gong, Zeyu and Tao, Bo},
  journal={IEEE Robotics and Automation Letters},
  year={2025},
  publisher={IEEE}
}

@article{xiang2025foldable,
  title={The Foldable Fin With Dynamic Adjustments for Manta Ray-Inspired Robots},
  author={Xiang, Yang and Gu, Le and Ye, Kangjie and Zhang, Zhenwei and Gong, Zeyu and Tao, Bo},
  journal={IEEE Robotics and Automation Letters},
  year={2025},
  publisher={IEEE}
}

@article{hao2024bioinspired,
  title={Bioinspired closed-loop CPG-based control of a robotic manta for autonomous swimming},
  author={Hao, Yiwei and Cao, Yonghui and Cao, Yingzhuo and Mo, Xiong and Huang, Qiaogao and Gong, Lei and Pan, Guang and Cao, Yong},
  journal={Journal of Bionic Engineering},
  volume={21},
  number={1},
  pages={177--191},
  year={2024},
  publisher={Springer}
}

@article{liu2022manta,
  title={A manta ray robot with soft material based flapping wing},
  author={Liu, Qimeng and Chen, Hao and Wang, Zhenhua and He, Qu and Chen, Linke and Li, Weikun and Li, Ruipeng and Cui, Weicheng},
  journal={Journal of Marine Science and Engineering},
  volume={10},
  number={7},
  pages={962},
  year={2022},
  publisher={MDPI}
}

@article{song2023development,
  title={Development of a biomimetic underwater robot for bottom inspection of marine structures},
  author={Song, Seokyong and Kim, Juhwan and Kim, Taesik and Song, Young-woon and Yu, Son-Cheol},
  journal={International Journal of Control, Automation and Systems},
  volume={21},
  number={12},
  pages={4041--4056},
  year={2023},
  publisher={Springer}
}

@article{odashima2002hierarchical,
  title={Hierarchical control structure of a multilegged robot for environmental adaptive locomotion},
  author={Odashima, Tadashi and Luo, Zhiwei and Hosoe, Shigeyuki},
  journal={Artificial Life and Robotics},
  volume={6},
  number={1},
  pages={44--51},
  year={2002},
  publisher={Springer}
}

@article{yuasa1990coordination,
  title={Coordination of many oscillators and generation of locomotory patterns},
  author={Yuasa, Hideo and Ito, M},
  journal={Biological Cybernetics},
  volume={63},
  number={3},
  pages={177--184},
  year={1990},
  publisher={Springer}
}

@inproceedings{zhangDynamicTargetTracking2024a,
  title = {Dynamic {{Target Tracking}} of {{Bionic Fish Based}} on {{Sliding Mode-CPG Controller}}},
  booktitle = {{{OCEANS}} 2024 - {{Singapore}}},
  author = {Zhang, Wei and Sun, Ruichi and Gong, Qingshuo and Han, Peiyu and Zhang, Zhe and Shi, Yefan},
  year = {2024},
  month = apr,
  pages = {1--6},
  urldate = {2025-07-16},
}

@article{struebig2020design,
  title={Design and development of the efficient anguilliform swimming robot—MAR},
  author={Struebig, Konstantin and Bayat, Behzad and Eckert, Peter and Looijestijn, Anouk and Lueth, Tim C and Ijspeert, Auke J},
  journal={Bioinspiration \& Biomimetics},
  volume={15},
  number={3},
  pages={035001},
  year={2020},
  publisher={IOP Publishing}
}

@article{tasadaDolphin,
  author={Asada, Takumi and Oki, Takao and Furuhashi, Hideo and Tabata, Kenta and Miyagusuku, Renato and Ozaki, Koichi},
  journal={IEEE Access}, 
  title={Performance Evaluation using Sim2Real of a Robotic Dolphin with Multi-link body Mechanism and CPG-based Controller}, 
  year={2025},
  volume={},
  number={},
  pages={1-1},
}

@article{li2024current,
  title={Current status and technical challenges in the development of biomimetic robotic fish-type submersible},
  author={Li, Jinyu and Li, Weikun and Liu, Qimeng and Luo, Bing and Cui, Weicheng},
  journal={Ocean-Land-Atmosphere Research},
  volume={3},
  pages={0036},
  year={2024},
  publisher={AAAS}
}

@article{cui2023review,
  title={Review of research and control technology of underwater bionic robots},
  author={Cui, Zhongao and Li, Liao and Wang, Yuhang and Zhong, Zhiwei and Li, Junyang},
  journal={Intelligent Marine Technology and Systems},
  volume={1},
  number={1},
  pages={7},
  year={2023},
  publisher={Springer}
}

@article{liu2024maneuverable,
  title={A maneuverable underwater vehicle for near-seabed observation},
  author={Liu, Kaixin and Ding, Mingxuan and Pan, Biye and Yu, Peiye and Lu, Dake and Chen, Siwen and Zhang, Shuo and Wang, Gang},
  journal={Nature Communications},
  volume={15},
  number={1},
  pages={10284},
  year={2024},
  publisher={Nature Publishing Group UK London}
}

\end{document}